\documentclass[sigconf]{acmart}
\AtBeginDocument{%
  \providecommand\BibTeX{{%
    \normalfont B\kern-0.5em{\scshape i\kern-0.25em b}\kern-0.8em\TeX}}}

\setcopyright{acmcopyright}
\copyrightyear{2018}
\acmYear{2020}
\acmConference[ACMMM 2020]{ACMMM 2020: ACM MULTIMEDIA 2020}{October 12-16, 2020}{Seattle, US}
\acmBooktitle{ACMMM 2020: ACM MULTIMEDIA,
  October 12-16, 2020, Seattle, US}
\acmSubmissionID{1246}

\usepackage{graphicx}
\usepackage{subfigure}
\makeatletter 
\renewcommand{\@thesubfigure}{\hskip\subfiglabelskip}
 \makeatother

\begin{document}

\newcommand{\hw}[1]{\textcolor{blue}{[Hongwei: #1]}}

%%
%% The "title" command has an optional parameter,
%% allowing the author to define a "short title" to be used in page headers.
\title[M$\mathbf{^3}$VSNet: Unsupervised Multi-metric Multi-view Stereo Network]{M$\mathbf{^3}$VSNet: Unsupervised Multi-metric\\ Multi-view Stereo Network}

%%
%% The "author" command and its associated commands are used to define
%% the authors and their affiliations.
%% Of note is the shared affiliation of the first two authors, and the
%% "authornote" and "authornotemark" commands
%% used to denote shared contribution to the research.
\author{Baichuan Huang, Hongwei Yi, Can Huang, Yijia He, Jingbin Liu, Xiao Liu}
%\authornote{Both authors contributed equally to this research.}
%\orcid{1234-5678-9012}
%\authornotemark[1]
%\email{webmaster@marysville-ohio.com}
\affiliation{%
  \institution{Wuhan University, Peking University, Megvii Technology Limited}
}

\renewcommand{\shortauthors}{Baichuan Huang, et al.}

%%
%% The abstract is a short summary of the work to be presented in the
%% article.
\begin{abstract}
The present Multi-view stereo (MVS) methods with supervised learning-based networks have an impressive performance comparing with traditional MVS methods. However, the ground-truth depth maps for training are hard to be obtained and are within limited kinds of scenarios. In this paper, we propose a novel unsupervised multi-metric MVS network, named M$\mathbf{^3}$VSNet, for dense point cloud reconstruction without any supervision. To improve the robustness and completeness of point cloud reconstruction, we propose a novel multi-metric loss function that combines pixel-wise and feature-wise loss function to learn the inherent constraints from different perspectives of matching correspondences. Besides, we also incorporate the normal-depth consistency in the 3D point cloud format to improve the accuracy and continuity of the estimated depth maps. Experimental results show that M$\mathbf{^3}$VSNet establishes the state-of-the-arts unsupervised method and achieves comparable performance with previous supervised MVSNet on the \textsl{DTU} dataset and demonstrates the powerful generalization ability on the \textsl{Tanks \& Temples} benchmark with effective improvement.

%The present Multi-view stereo (MVS) methods with supervised learning-based networks have an impressive performance comparing with traditional MVS methods. However, the ground-truth depths for training are hard to be obtained and are limited kinds of scenarios. In this paper, we propose a novel unsupervised multi-metric MVS network, named M3VSNet, for dense point cloud reconstruction which does not need ground-truth for training. To improve the robustness and completeness of point cloud reconstruction in textureless area, we propose a novel multi-metric loss function that combines pixel-wise and feature-wise loss function to learn the inherent constraints from different perspectives of matching correspondences. Besides, we also incorporate the normal-depth consistency in the 3D point cloud format to improve the accuracy and continuity of the estimated depth maps. Experimental results show that M3VSNet establishes the state-of-the-arts unsupervised method and achieves comparable performance with original MVSNet on the DTU dataset and demonstrates the powerful generalization ability on the Tanks & Temples benchmark with effective improvement. Our code is available at \href{https://github.com/whubaichuan/M3VSNet}{https://github.com/whubaichuan/M3VSNet.}

\end{abstract}

%%
%% The code below is generated by the tool at http://dl.acm.org/ccs.cfm.
%% Please copy and paste the code instead of the example below.
%%



%%
%% Keywords. The author(s) should pick words that accurately describe
%% the work being presented. Separate the keywords with commas.
%\keywords{Multi-view stereo, Unsupervised network, Multi-metric}

%% A "teaser" image appears between the author and affiliation
%% information and the body of the document, and typically spans the
%% page.
%\begin{teaserfigure}
%  \includegraphics[width=\textwidth]{sampleteaser}
%  \caption{Seattle Mariners at Spring Training, 2010.}
%  \Description{Enjoying the baseball game from the third-baseseats. Ichiro Suzuki preparing to bat.}
%  \label{fig:teaser}
%\end{teaserfigure}

%%
%% This command processes the author and affiliation and title
%% information and builds the first part of the formatted document.
\maketitle

\section{Introduction}
Multi-view stereo (MVS) aims to reconstruct the 3D dense point cloud from multi-view images \cite{cui2017hsfm,frahm2010building}, which has various applications in augmented reality, virtual reality and robotics, etc. \cite{seitz2006comparison}. 
Big progress has been made in the dense reconstruction with traditional methods through the hand-crafted features (e.g. NCC) to calculate the matching correspondences \cite{Furukawa2007AccurateDA,goesele2006multi,schonberger2016pixelwise,galliani2015massively,vu2011high}. Though, the efficient and robust methods of MVS in the large-scale environments are still the challenging tasks \cite{seitz2006comparison}. Recently, deep learning is introduced to relieve this limitation. The supervised learning-based MVS methods achieve significant progress especially improving the efficiency and completeness of dense point cloud reconstruction \cite{ji2017surfacenet,ummenhofer2017demon,yao2018mvsnet,yao2019recurrent}. These learning-based methods learn and infer the information to handle matching ambiguity which is hard to be obtained by stereo correspondences. However, these supervised learning-based methods strongly depend on the training datasets with ground-truth depth maps, which have limited kinds of scenarios and are not easy to be available. Thus it is a big hurdle and may lead to bad generalization ability in different complex scenarios \cite{dai2019mvs2,gu2019cascade,yao2018mvsnet,yang2018unsupervised}. Furthermore, the robustness and completeness of dense point cloud reconstruction still have a lot of room to be improved. Current unsupervised learning-based methods are mainly based on the pixel-wise level, which will cause incorrect matching correspondences with low robustness \cite{yang2018unsupervised}\cite{yao2018mvsnet}. Because for two identical images, the difference will be huge as long as pixel offset from the perspective of pixel level. However, they are almost the same from the perspective of perception such as feature level. In addition, the human visual system perceives the surrounding world depending on the object features rather than a single image pixel \cite{benzhang2018unsupervised}. 
Therefore, in order to improve the robustness and completeness of unsupervised learning-based MVS, it drives us to consider the similarity on object features.

In this paper $\footnote{Can Huang is the corresponding author}$, we propose a novel unsupervised multi-metric MVS network, named M$\mathbf{^3}$VSNet as shown in figure \ref{fig:network}, which could infer the depth maps for dense point cloud reconstruction even in non-ideal environments. Most importantly, we propose a novel multi-metric loss function, namely pixel-wise and feature-wise loss function. The key insight is that the human visual system perceives the surrounding world by the object features \cite{benzhang2018unsupervised}. In terms of this loss function, both the photometric and geometric matching consistency can be well guaranteed, which is more accurate and robust compared with the only photometric constraints used in MVSNet \cite{yao2018mvsnet}. Specifically, we introduce the multi-scale feature maps from the pre-trained VGG16 network as vital clues in the feature-wise loss. Low-level feature representations learn more texture details while high-level features learn semantic information with a large receptive field. Different level features are the representations of different receptive fields. By aggregating multi-scale features, our proposed M$\mathbf{^3}$VSNet can consider both the low-level image texture and the high-level semantic information. Therefore, the network can well improve the robustness and accuracy of matching correspondences. Compared with the network only using pixel-wise loss which performs mismatch errors in some challenging scenarios such as textureless, mirror effect or reflection and texture repeat areas \cite{seitz2006comparison,yang2018unsupervised,yao2018mvsnet}, M$\mathbf{^3}$VSNet can improve the robustness by considering the similarity between the multi-scale semantic features.

Besides, in order to further improve the performance of the estimated depth maps, we incorporate the normal-depth consistency in the world coordinate space to constraint the local surface tangent obtained from the estimated depth maps to be orthogonal to the calculated normal. This regularization will improve the accuracy and continuity of the estimated depth maps. Moreover, we utilize the multi-scale pyramid feature aggregation to construct the 3D cost volume with more contextual information to improve the robustness and accuracy of feature correspondences.

Our main contributions are summarized as below:
\begin{itemize}
\item We propose a novel multi-metric unsupervised network, which can work even in non-ideal environments, for multi-view stereo without any ground-truth 3D training data.
\item we propose a novel multi-metric loss function that considers different perspectives of matching correspondences beyond pixel value. Besides, we incorporate the normal-depth consistency in the 3D point cloud format to improve the accuracy and continuity of the estimated depth maps.  
\item Extensive experiments demonstrate that our proposed M$\mathbf{^3}$VSNet outperforms the previous state-of-the-art unsupervised methods and achieves comparable performance with original MVSNet on the \textsl{DTU} dataset and shows the excellent generalization ability on the \textsl{Tanks \& Temples} benchmark with effective improvement.
\end{itemize}

\section{Related Work}
\subsection{Traditional MVS}

Many traditional methods have been proposed in this field such as voxel-based method \cite{sinha2007multi}, feature points diffusion \cite{Furukawa2007AccurateDA} and the fusion of estimated depth maps \cite{barnes2009patchmatch}. First of all, the voxel-based method consumes many computing resources and its accuracy depends on the resolution of voxel mainly \cite{ji2017surfacenet}. Secondly, the blank area may seriously suffer from the textureless problem in the method of feature points diffusion. Thirdly, the most used method is the fusion of inferred depth maps, which gets the depth maps and then fuses all the depth maps together to the final point cloud \cite{campbell2008using}. Besides, many methods of improvement have been proposed. Silvano \cite{galliani2015massively} formulates the patch matches in 3D space and the progress can be massively parallelized and delivered. Johannes \cite{schonberger2016pixelwise} estimates the depth and normal maps synchronously and uses photometric and geometric priors to refine the image-based depth and normal fusion. Though, the robustness needs to be improved when dealing with non-ideal environments such as textureless or texture repeat areas and no-Lambert surfaces.

\subsection{Depth Estimation}

The fusion of estimated depth maps can decouple the reconstruction into depth estimation and depth fusion. Depth estimation with monocular video and binocular image pairs has many similarities with the multi-view stereo here \cite{laga2019survey}. But there are exactly some differences between them. Monocular video \cite{zhou2017unsupervised} lacks the real scale of the depth actually and binocular image pairs always need to rectify the parallel two images \cite{dosovitskiy2015flownet}. In this case, only the disparity needs to be inferred without considering the intrinsic and extrinsic of the camera. As for multi-view stereo, the input is the arbitrary number of pictures. What's more, the transformation among these positions should be taken into consideration as a whole \cite{yao2018mvsnet}. Other obstacles such as multi-view occlusion and consistency \cite{dai2019mvs2} raise the bar for depth estimation of multi-view stereo than that of monocular video and binocular image pairs.

\subsection{Supervised Learning MVS}
Since Yao Yao proposed MVSNet in 2018 \cite{yao2018mvsnet}, many supervised networks based on MVSNet have been proposed. To reduce GPU memory consumption, Yao Yao introduces R-MVSNet with the help of gated recurrent unit \cite{yao2019recurrent}. Gu uses the concept of the cascade to shrink the cost volume \cite{gu2019cascade}. Yi introduced two new self-adaptive view aggregation with pyramid multi-scale images to enhance the point cloud in textureless regions \cite{yi2019pyramid}. Luo utilizes the plane-sweep volumes with isotropic and anisotropic 3D convolutions to get better results \cite{luo2019p}. In this kind of task, cost volume and 3D regularization are highly memory-consuming. More importantly, the ground-truth depth maps are derived from heavy labor.

\subsection{Unsupervised Learning MVS}
The unsupervised network utilizes the photometric and geometric constraints to learn the depth by itself, which relief the complicated artificial markers for ground-truth depth maps. Many works explore unsupervised learning in monocular video and binocular image pairs. Reza \cite{mahjourian2018unsupervised} presents the unsupervised learning method for depth and ego-motion from monocular video. The paper uses image reconstruction loss, 3D point cloud alignment loss and additional image-based loss. Being similar to unsupervised learning in monocular video and binocular image pairs \cite{alhashim2018high}, the losses of MVS are also based on photometric and geometric consistency. Dai \cite{dai2019mvs2} predicts the depth maps for all views simultaneously in a symmetric way. In the stage, cross-view photometric and geometric consistency can be guaranteed. But this method consumes a lot of GPU memory. Additionally, Tejas \cite{khot2019learning} proposes the simplified network and traditional loss designation but an unsatisfied result. Efforts are worthy to be paid in this direction.

\section{M$\mathbf{^3}$VSNet}
In this section, we introduce our proposed M$\mathbf{^3}$VSNet in detail. We firstly describe the network architecture in section \ref{sec:network_arch} to generate initial depth map, then illustrate the normal-depth consistency in section \ref{sec:nd_consistency} to refine it in consideration of the orthogonality between normal and local surface tangent. Finally, our proposed novel multi-metric loss in section \ref{sec:multi-metric} is introduced by considering different perspectives of matching correspondences to improve the robustness and completeness of point cloud reconstruction.

\subsection{Network Architecture}
\label{sec:network_arch}
\begin{figure*}[t]
\begin{center}
\includegraphics[width=1.0\linewidth]{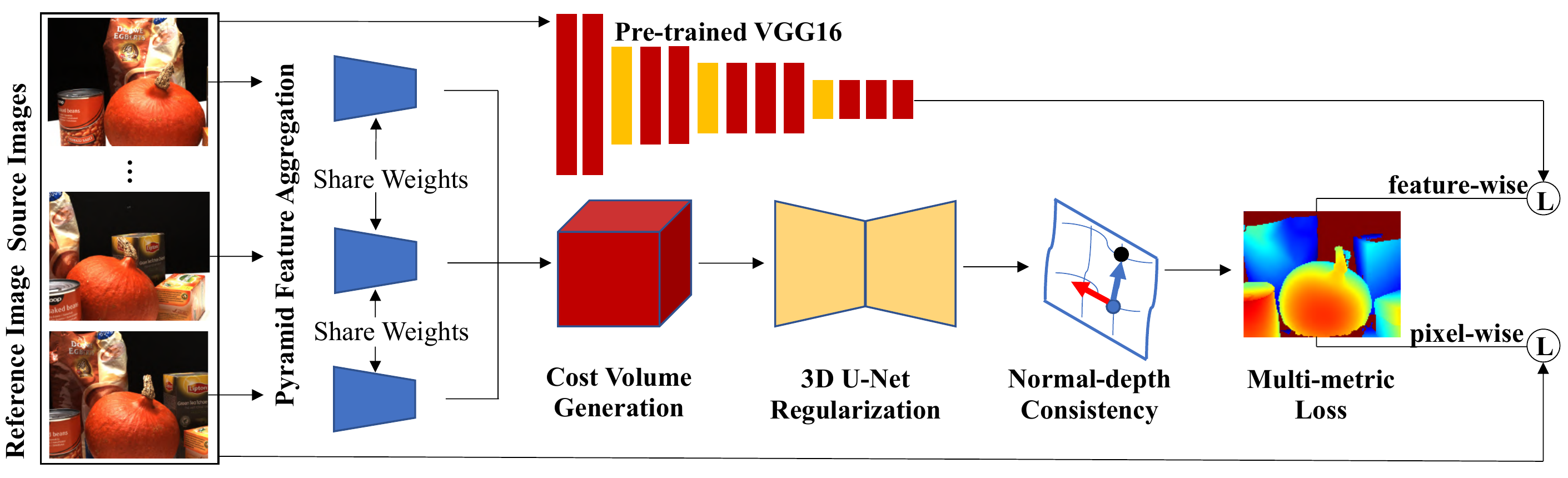}
\end{center}
   \caption{The architecture of our proposed M$\mathbf{^3}$VSNet. It contains five components: pyramid feature aggregation, variance-based cost volume generation, 3D U-Net regularization, normal-depth consistency and multi-metric loss function.}
\label{fig:network}
\end{figure*}

The basic architecture of our proposed M$\mathbf{^3}$VSNet consists of three parts, namely pyramid feature aggregation, variance-based cost volume generation and 3D U-Net regularization, as shown in figure \ref{fig:network}. The pyramid feature aggregation extracts features from low-level to high-level representations with more contextual information. Then the same variance-based cost volume generation and 3D U-Net regularization as MVSNet \cite{yao2018mvsnet} are used to generate the initial depth map. The advance architecture of M$\mathbf{^3}$VSNet consists of two parts, namely normal-depth consistency and multi-metric loss. After generating the initial depth map, we incorporate the novel normal-depth consistency to refine it in consideration of the orthogonality between normal and local surface tangent. More importantly, we construct multi-metric loss, which consists of pixel-wise loss and feature-wise loss. We will briefly describe each module in the following parts.
\subsubsection{Pyramid Feature Aggregation}
\label{sec:pyramid}

\begin{figure}[t]
\begin{center}
\includegraphics[width=0.7\linewidth]{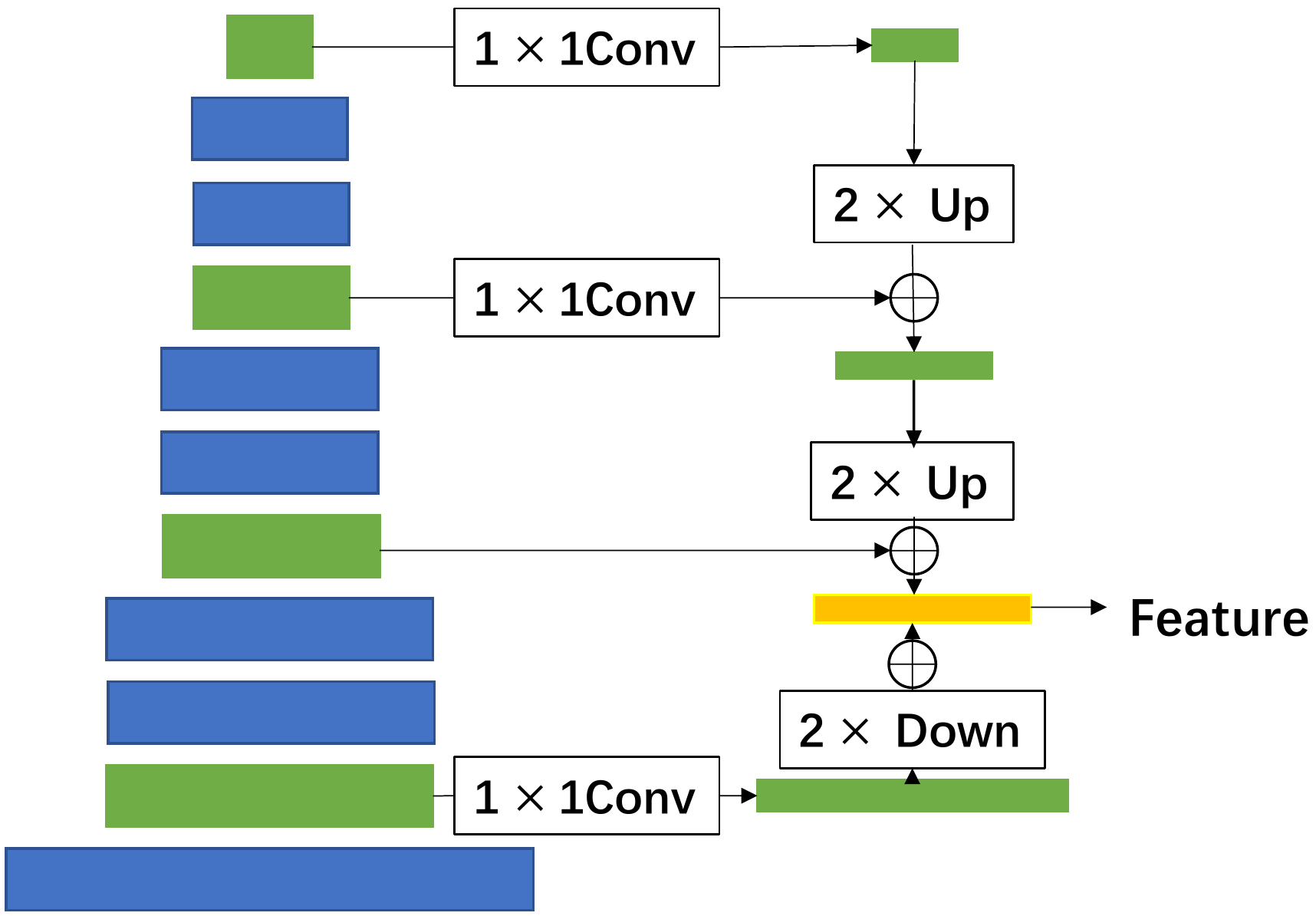}
\end{center}
   \caption{The illustration of pyramid feature aggregation.}
\label{fig:feature}
\end{figure}

In previous supervised learning-based network such as MVSNet \cite{yao2018mvsnet}, only the 1/4 feature is adopted (1/4 represents a quarter of the size of the original reference image). 1/4 feature lacks multi-scale context information for matching correspondences. Therefore, we propose to use the pyramid feature aggregation that aggregates different scale features with contextual information of different receptive fields \cite{lin2017feature}. Figure \ref{fig:feature} shows the details of this module. For the input images, the feature extraction network is constructed to extract the aggregated 1/4 feature. In the process of bottom-up, the stride of the layer 3, 6 and 9 is set to 2 to get the four scale features in eleven-layer 2D CNN. Each convolutional layer is followed by the structure of BatchNorm and ReLU. In the process of up-bottom, each level of features is derived from the concatenate by the upsampling of the higher layer and the feature in the same layer with fewer channels. Especially, the 1/2 feature needs to be downsampled to be aggregated into the final 1/4 feature. To reduce the dimension of the final 1/4 feature, the $1\times1$ convolution for each concatenation is adopted. At last, we get the final feature with 32 channels, which is an aggregation of contextual information from low-level to high-level representations.

\subsubsection{Cost volume and 3D U-Net regularization}

The construction of variance-based cost volume is based on the differentiable homography warping with the number of different depth hypotheses $D$ in MVSNet \cite{yao2018mvsnet}. Then 3D U-Net regularization is used to regularize the 3D cost volume, which is simple but effective for aggregating features. At last, the initial depth is derived from the $soft$ $argmin$ operation with the probability volume after the regularization.

\subsection{Normal-depth Consistency}
\label{sec:nd_consistency}

The initial depth still contains some incorrect matching correspondences. Therefore, to improve the quality of estimated depth maps further, we incorporate the normal-depth consistency based on the orthogonality between normal and local surface tangent \cite{yang2018unsupervised}. The consistency will make the depth more reasonable in 3D space. Normal-depth consistency can be divided into two steps. Firstly, the normal should be calculated by the depth with the orthogonality. Then the refined depth can be inferred by the normal and initial depth according to the projection relationship. This module cooperating with 3D U-Net will refine the depth in 2D and 3D space jointly, which improves the accuracy and continuity of depth.  

\begin{figure}[t]
\begin{center}
\includegraphics[width=1.0\linewidth]{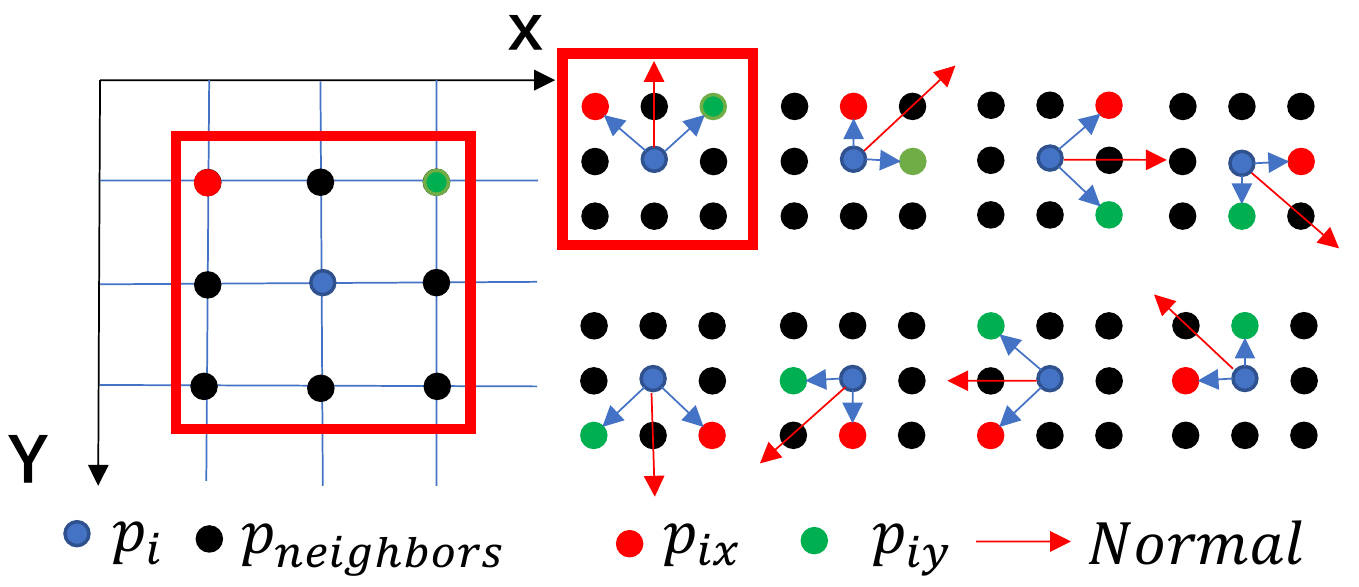}
\end{center}
   \caption{The illustration of normal from depth}
\label{fig:normal}
\end{figure}

As shown in figure \ref{fig:normal}, eight neighbors are selected to infer the normal of the central point. Due to the orthogonality, the operation of cross-product is used. For each central point $p_i$, one set of the neighbors can be recognized as $p_{ix}$ and $p_{iy}$. If the depth $Z_i$ of $p_i$ and the intrinsics $K$ of camera are known, the normal $\widetilde{N_i}$ can be calculated as below:
\begin{equation}
P_i=K^{-1}Z_ip_i
\end{equation}
\begin{equation}
\widetilde{N_i}=\overrightarrow{P_iP_{ix}}\times\overrightarrow{P_iP_{iy}}
\end{equation}
To add the credibility of final normal estimation ${N_i}$, mean cross-product for eight neighbors can be presented as below:
\begin{equation}
N_i=\frac{1}{8}\sum_{i=1}^{8}(\widetilde{N_i})
\end{equation}

\begin{figure}[t]
\begin{center}
\includegraphics[width=0.9\linewidth]{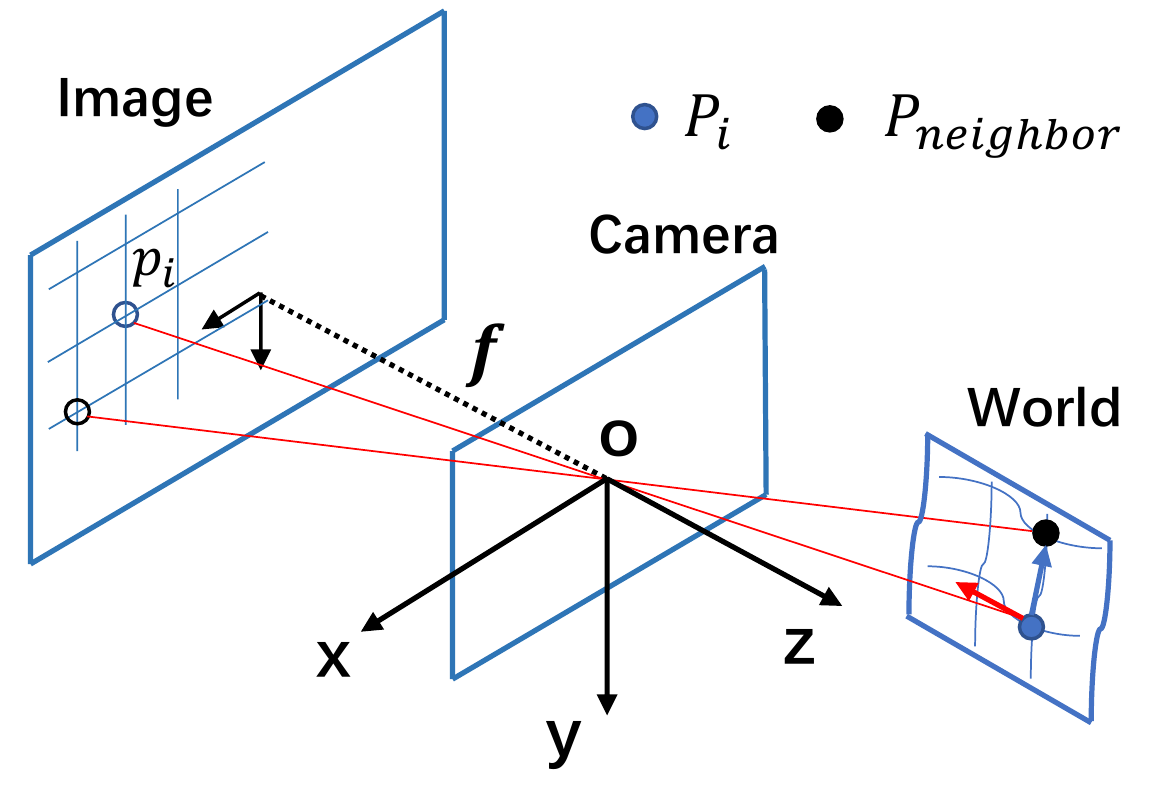}
\end{center}
   \caption{The illustration of depth from normal}
\label{fig:depth}
\end{figure}

The final refined depth maps can be available when the normal and initial depth maps are provided. In figure \ref{fig:depth}, for each pixel $p_i(x_i,y_i)$, the depth of the neighbor $p_{neighbor}$ should be refined. Their corresponding 3D points are $P_i$ and $P_{neighbor}$. The normal of $P_i$ is $\overrightarrow{N_i}(n_x,n_y,n_z)$. The depth of $P_i$ is $Z_i$ and the depth of $P_{neighbor}$ is $Z_{neighbor}$. We can get the equation $\overrightarrow{N}\perp \overrightarrow{P_iP_{neighbor}}$. The relationship is apparently reasonable due to the orthogonality and surface consistency in the local surface. In summary, the depth $Z_{neighbor}$ of the neighbors can be inferred by the depth and normal of the central point.

\begin{equation}
(K^{-1}Z_ip_i-K^{-1}Z_{neighbor}p_{neighbor})\begin{bmatrix}
n_x\\ 
n_y\\ 
n_z
\end{bmatrix}=0
\end{equation}

For the refined depth, eight neighbors are also taken into consideration. Considering the discontinuity of normal in some edge or irregular surface, the weight $w_{i}$ for the reference image $I_i$ is introduced to make depth more conforming to geometric consistency. The weight is defined as below:

\begin{equation}
w_{i}=e^{-\alpha_1\left | \bigtriangledown I_i \right |}
\end{equation}

The weight $w_{i}$ depends on the gradient between $p_i$ and $p_{neighbor}$, which means that the bigger gradient represents the less reliability of the refined depth. In view of the eight neighbors, the final refined depth $\widetilde Z_{neighbor}$ is a combination of the weighted sum of eight different directions. The final refined depth is the result of regularization in 3D space, which improves the accuracy and continuity of the estimated depth maps.
\begin{equation}
\widetilde Z_{neighbor}=\sum_{i=1}^{8}w_{i}'Z_{neighbor}
\end{equation}
\begin{equation}
w_{i}'==\frac{w_{i}}{\sum_{i=1}^{8}w_{i}}
\end{equation}

\subsection{Multi-metric Loss}
\label{sec:multi-metric}
We propose a novel multi-metric loss function by considering different perspectives of matching in feature correspondence beyond pixel, which is quite crucial and effective. The pixel-wise loss can guarantee the matching correspondences with more texture details and the feature-wise loss can make use of the semantic information. 

The key idea embodied in multi-metric loss function is the photometric consistency crossing multi-views \cite{barnes2009patchmatch}. Given the reference image $I_{ref}$ and source image $I_{src}$, the corresponding intrinsic parameters are represented as $K_{ref}$ and $K_{src}$. Besides, the extrinsic from $I_{ref}$ to $I_{src}$ is represented as $T$. For the pixel $p_i(x_i,y_i)$ in $I_{ref}$, the corresponding pixel $p'_i(x'_i,y'_i)$ in $I_{src}$ can be calculated as:

\begin{equation}
p'_i=KT( K^{-1}\widetilde Z_{i}p_i)
\end{equation}

The overlapping area, named $I'_{src}$, from reference image $I_{ref}$ to source image $I_{src}$ can be sampled using the 	bilinear interpolation. 
\begin{equation}
I'_{src}=I_{src}(p'_i)
\end{equation}

For the occlusion area, the value of the pixel in $I'_{src}$ is set to zero. Obviously, the mask $M$ can be obtained when the $p_i$ is projected to the external area of $I_{src}$. Based on the prior constraint, the multi-metric loss function $L$ is formulated as the sum of pixel-wise loss $L_{pixel}$ and feature-wise loss $L_{feature}$.

\begin{equation}
L=\sum (\gamma_1 L_{pixel}+\gamma_2  L_{feature})
\end{equation}

\subsubsection{Pixel-wise Loss}

\label{sec:loss_base}
For the pixel-wise loss, we only consider the photometric consistency between the reference image $I_{ref}$ and other source images.
There are mainly three parts of this loss function. Firstly, the photometric loss compares the difference of pixel value between $I_{ref}$ and $I'_{src}$. To relieve the influence of lighting changes, the gradient of every pixel is integrated into $L_{photo}$. 

\begin{equation}
L_{photo}=\frac{1}{m}\sum ((I_{ref}-I'_{src})+(\bigtriangledown I_{ref}-\bigtriangledown I'_{src})) \cdot M
\end{equation}
Where $m$ is the sum number of valid points in the mask $M$.

Secondly, the loss of structure similarity (SSIM) $L_{SSIM}$ is set to measure the similarity between $I_{ref}$ and $I'_{src}$. The operation $S$ will be $1$ when $I_{ref}$ is the same as $I'_{src}$.
\begin{equation}
L_{SSIM}=\frac{1}{m}\sum  \frac{1-S(I_{ref},I'_{src})}{2}   \cdot M
\end{equation}

Thirdly, the smooth of final refined depth map can 
make it less steep in the first-order domain and the second-order domain.
\begin{equation}
L_{smooth}=\frac{1}{n}\sum (e^{-\alpha_2\left | \bigtriangledown I_{ref} \right |}\left |  \bigtriangledown \widetilde Z_{i} \right |+ e^{-\alpha_3\left | \bigtriangledown^2  I_{ref} \right |}\left |  \bigtriangledown^2 \widetilde Z_{i} \right |)
\end{equation}
Where $n$ is the sum number of points in reference image $I_{ref}$.

Finally, the total pixel-wise loss $L_{pixel}$ can be illustrated as below:
\begin{equation}
L_{pixel}=\lambda_1 L_{photo}+\lambda_2 L_{SSIM}+\lambda_3 L_{smooth}
\end{equation}

\subsubsection{Feature-wise Loss}

The network only using pixel-wise loss performs mismatch errors in some challenging scenarios such as textureless and texture repeat areas. In addition to the pixel-wise loss, one of the main improvements of M$\mathbf{^3}$VSNet is the use of feature-wise loss. Just like in image style transfer, perceptual loss combining with per-pixel loss improves the performance of style transfer in quality \cite{johnson2016perceptual}. The feature-wise loss will utilize more semantic information for matching correspondences.

\begin{figure}[t]
\begin{center}
\includegraphics[width=1.0\linewidth]{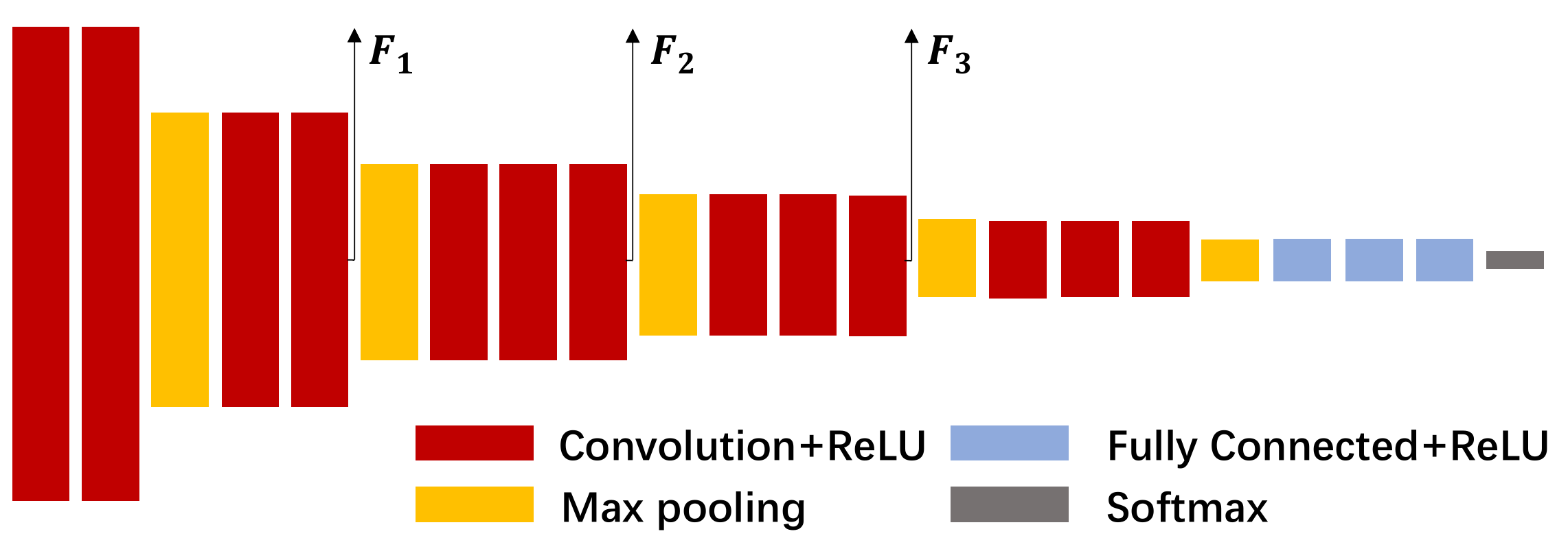}
\end{center}
   \caption{Feature-wise extraction from pre-trained VGG16}
\label{fig:vgg}
\end{figure}

Due to the strong correlation between the estimated depth and pyramid feature network mentioned in section \ref{sec:pyramid}, the high-level feature is extracted from pre-trained VGG16 instead of the pyramid feature network. Through the pre-trained VGG16 network, shown in figure \ref{fig:vgg}, the reference image $I_{ref}$ can extract more semantic high-level information to construct the feature-wise loss function. Here, we extract the layer 8, 15 and 22, which are one half, a quarter and one-eighth the size of the original input images. As a matter of fact, layer 3 output the same size of the original input image, which is actually the reuse of pixel-wise loss function. 

For every feature from the VGG16, we construct the loss based on the concept of crossing multi-views. Being similar to section \ref{sec:loss_base}, the corresponding pixel $p'_i$ in $F_{src}$ can be available. The matching features from $F_{ref}$ to $F_{src}$ can be presented as below:

\begin{equation}
F'_{src}=F_{src}(p'_i)
\end{equation}

The feature domain has a bigger receptive field, which is inspired by that human visual system perceives the scene by its features rather than a single pixel. Therefore, the obstacle of non-ideal areas can be relieved to some extent. The estimated final depth will detect the similarity of features beyond pixel texture value, which benefits from semantic information. The loss $L_{F}$ is represented as below:

\begin{equation}
L_{F}=\frac{1}{m}\sum (F_{ref}-F'_{src}) \cdot M
\end{equation}

The final feature-wise loss function is a weighted sum of different scale of features, which raises the robustness and completeness of point cloud reconstruction. $L_{F_8}$ represents the feature of layer 8 from pre-trained VGG16.

\begin{equation}
L_{feature}=\beta_1 L_{F_8}+\beta_2 L_{F_{15}}+\beta_3 L_{F_{22}}
\end{equation}

\section{Experiments}
\label{sec:experiments}
We conduct abundant experiments of our proposed M$\mathbf{^3}$VSNet on different datasets. Firstly, we evaluate M$\mathbf{^3}$VSNet on the \textsl{DTU} dataset and our method outperforms all the previous unsupervised MVS network \cite{dai2019mvs2,khot2019learning}. Then the ablation studies are carried out to find out potential improvements from our proposed different modules in section \ref{sec:ablation}. At last, we test M$\mathbf{^3}$VSNet on the \textsl{Tanks and Temples} benchmark to verify the generalization ability of our model.

\subsection{Performance on \textsl{DTU}}
The DTU dataset is a multi-view stereo dataset that has 124 different scenes with 49 scans for each scene, which is collected by the robotic arms \cite{jensen2014large}\cite{Aans2016LargeScaleDF}. With the lighting change, each scan has seven conditions with the known pose. We use the same train-validation-test split as in MVSNet \cite{yao2018mvsnet} and $\mathrm{MVS^2}$ \cite{dai2019mvs2}. That is to say, the scenes 1, 4, 9, 10, 11, 12, 13, 15, 23, 24, 29, 32, 33, 34, 48, 49, 62, 75, 77, 110, 114, 118 are selected as the test lists.

\subsubsection{Implementation Detail}
M$\mathbf{^3}$VSNet is implemented by Pytorch \cite{steiner2019pytorch:}. During the training phase, we only use the \textsl{DTU}'s training set without any  ground-truth depth maps. The resolution of the input image is the crop version of the original picture. That is 640 $\times$ 512. Due to the pyramid feature aggregation, the resolution of the final depth is 160 $\times$ 128. Additionally, the hypothetical range of depth is sampled from 425mm to 935mm and the depth sample number $D$ is set to 192. The model is trained with the batchsize as 4 in four NVIDIA RTX 2080Ti. By the pattern of data-parallel, each GPU with around 11G available memory could deal with the multi-batch. By using adam optimizer for 10 epochs, the learning rates are set to 1e-3 for the first epoch and decrease by 0.5 for every two epochs. For the balance of different weights in loss, we set $\gamma_1=1$, $\gamma _2=1$, $\alpha_1=0.1$, $\alpha_2=0.5$, $\alpha_3=0.5$, $\lambda_1=0.8$, $\lambda_2=0.2$, $\lambda_3=0.067$. Beyond that, $\beta_1=0.2$, $\beta_2=0.8$, $\beta_3=0.4$. During each iteration, one reference image and two source images are used. During the testing phase, the resolution of input image is 1600 $\times$ 1200.

\subsubsection{Results on \textsl{DTU}}

\begin{figure*}[t]
\centering
\subfigure{
\includegraphics[width=0.2\linewidth]{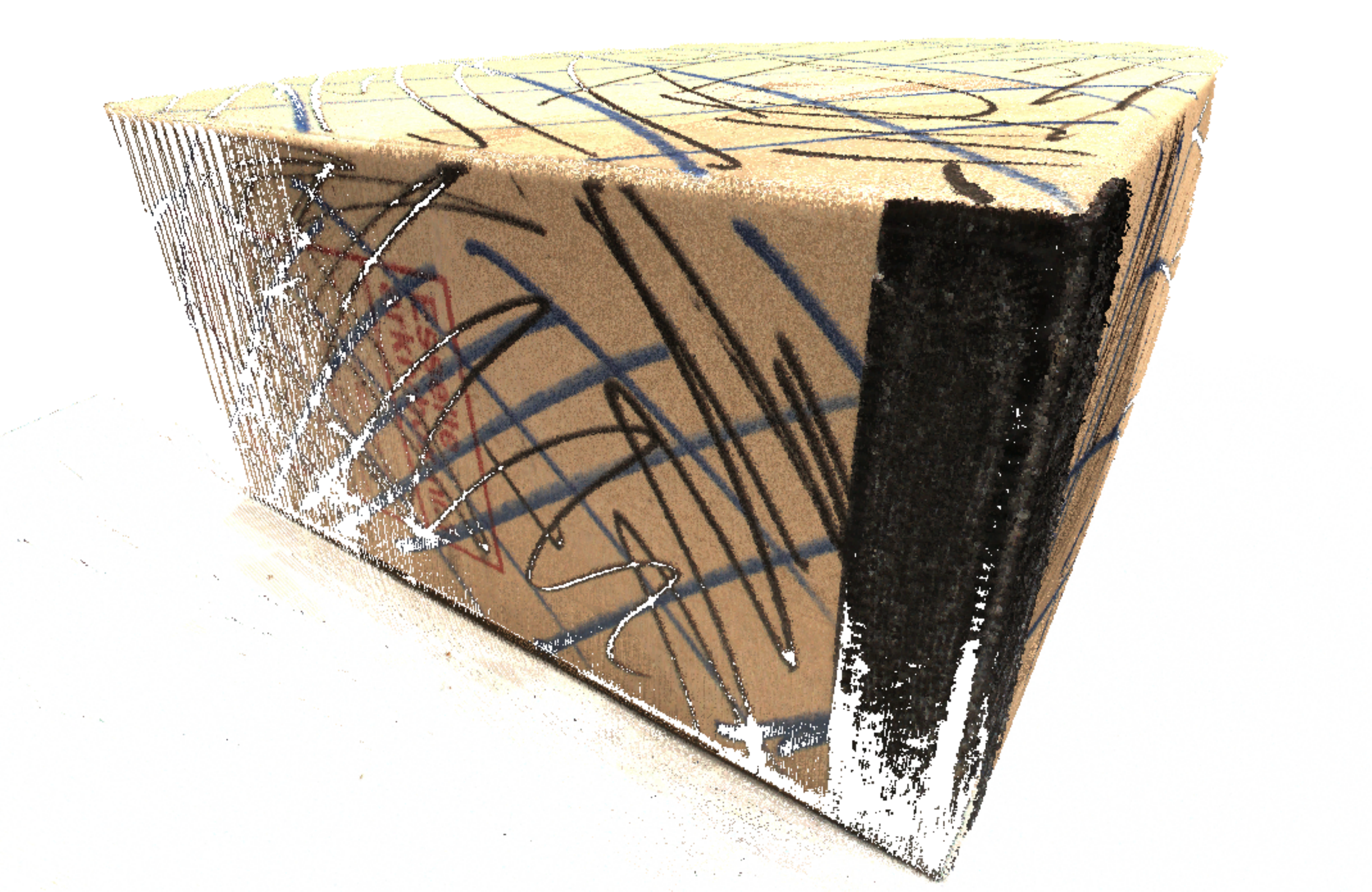}
}
\quad
\subfigure{
\includegraphics[width=0.2\linewidth]{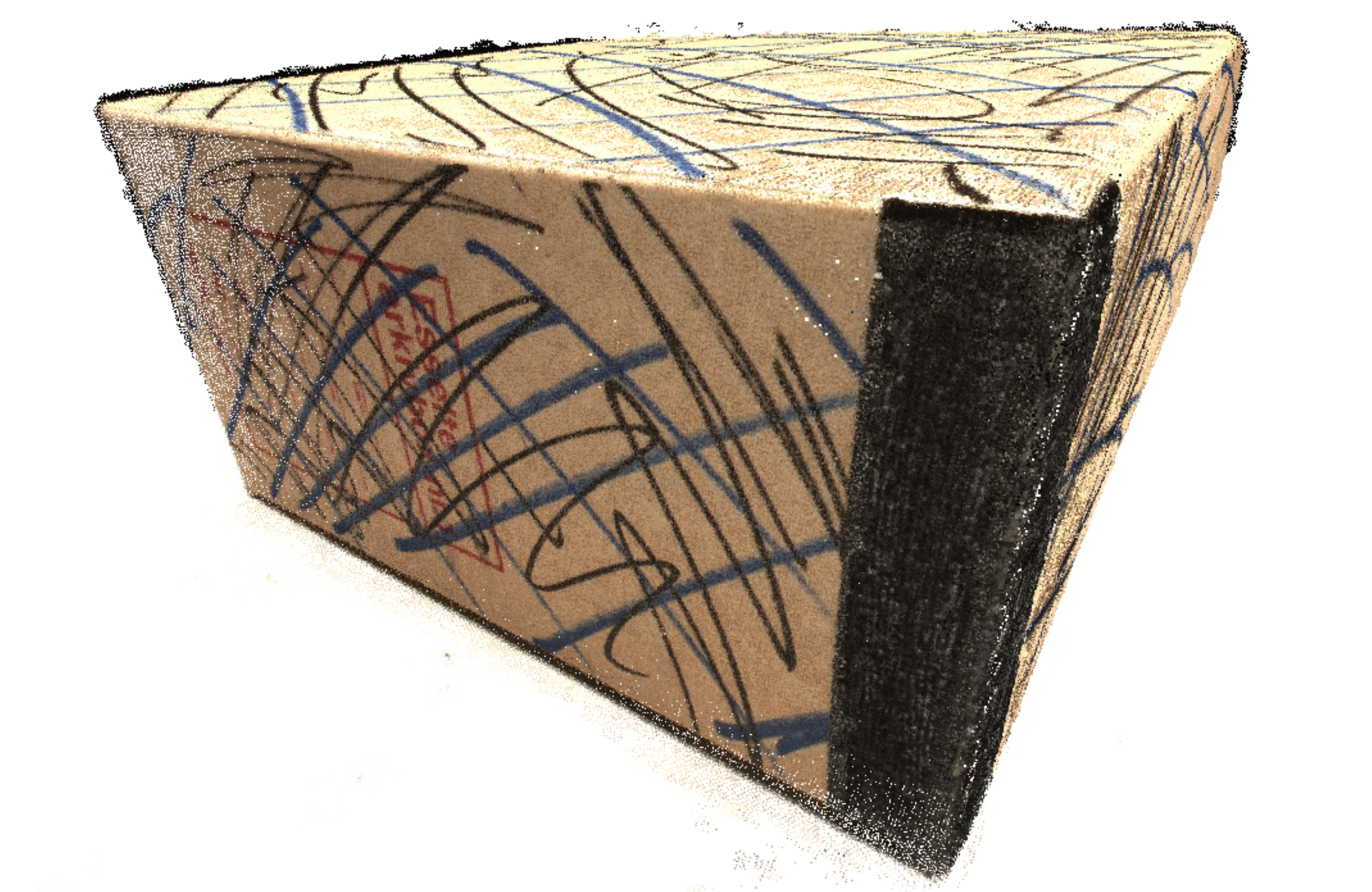}
}
\quad
\subfigure{
\includegraphics[width=0.2\linewidth]{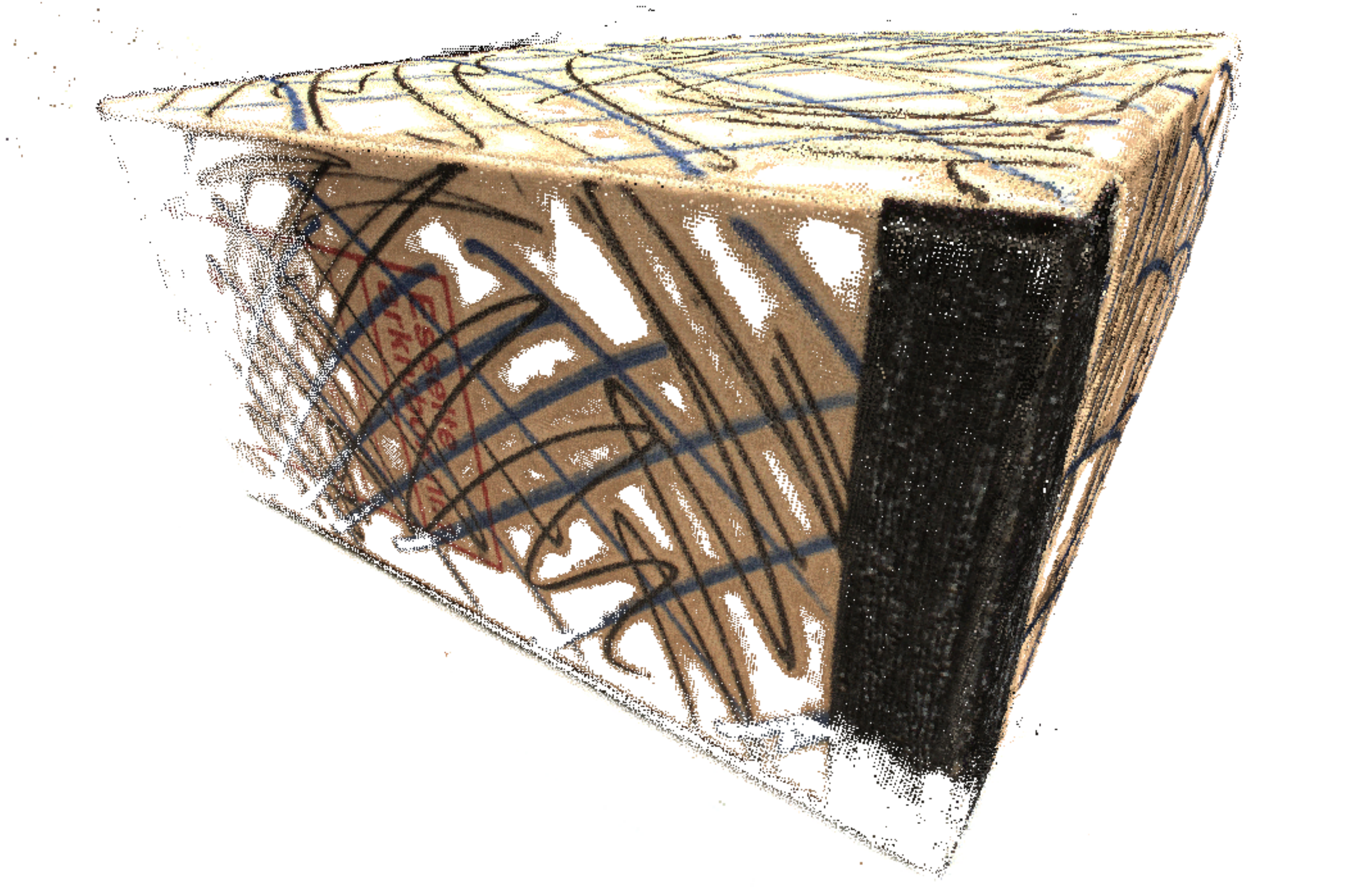}
}
\quad
\subfigure{
\includegraphics[width=0.2\linewidth]{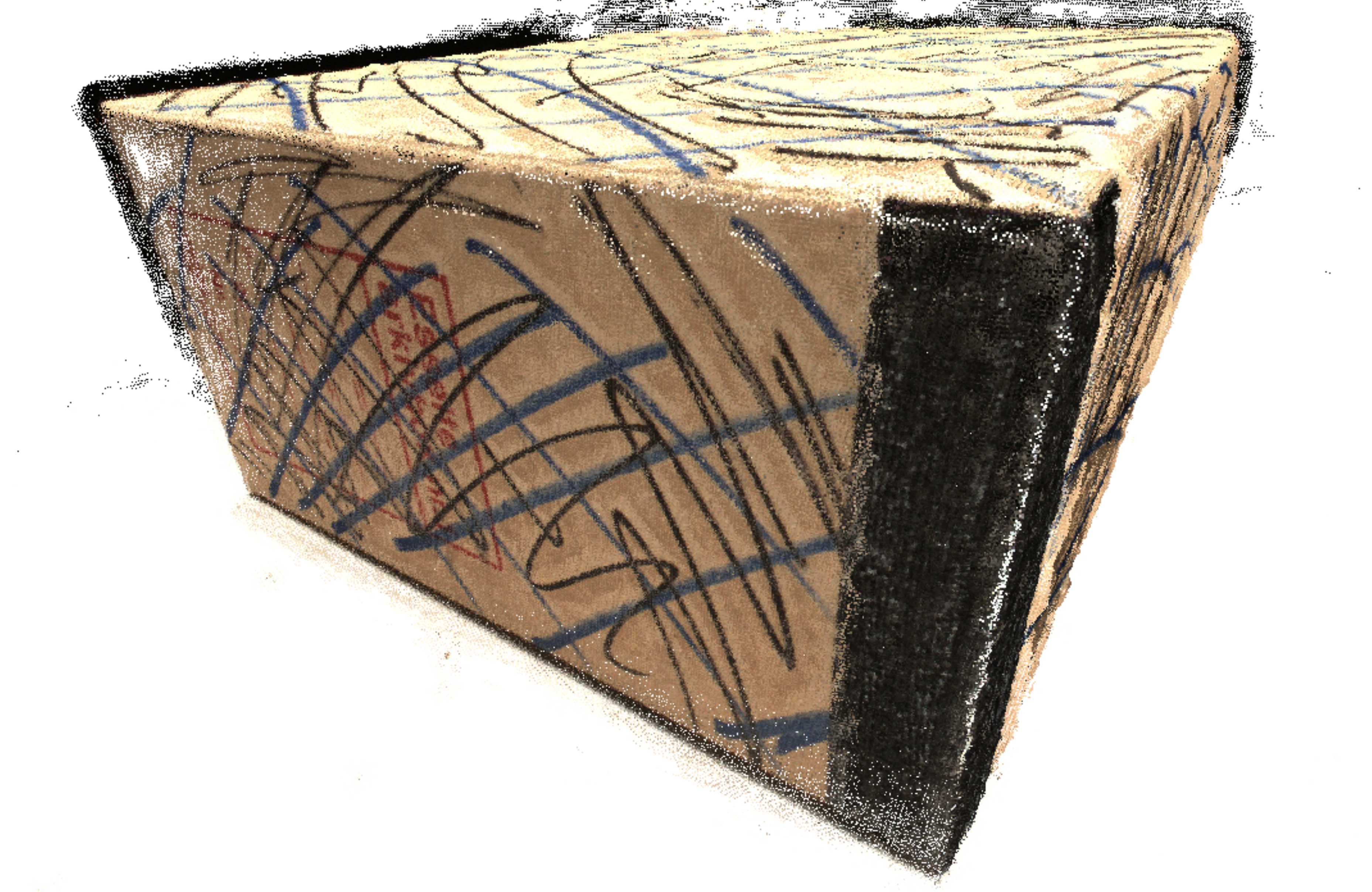}
}
\quad
\subfigure{
\includegraphics[width=0.2\linewidth]{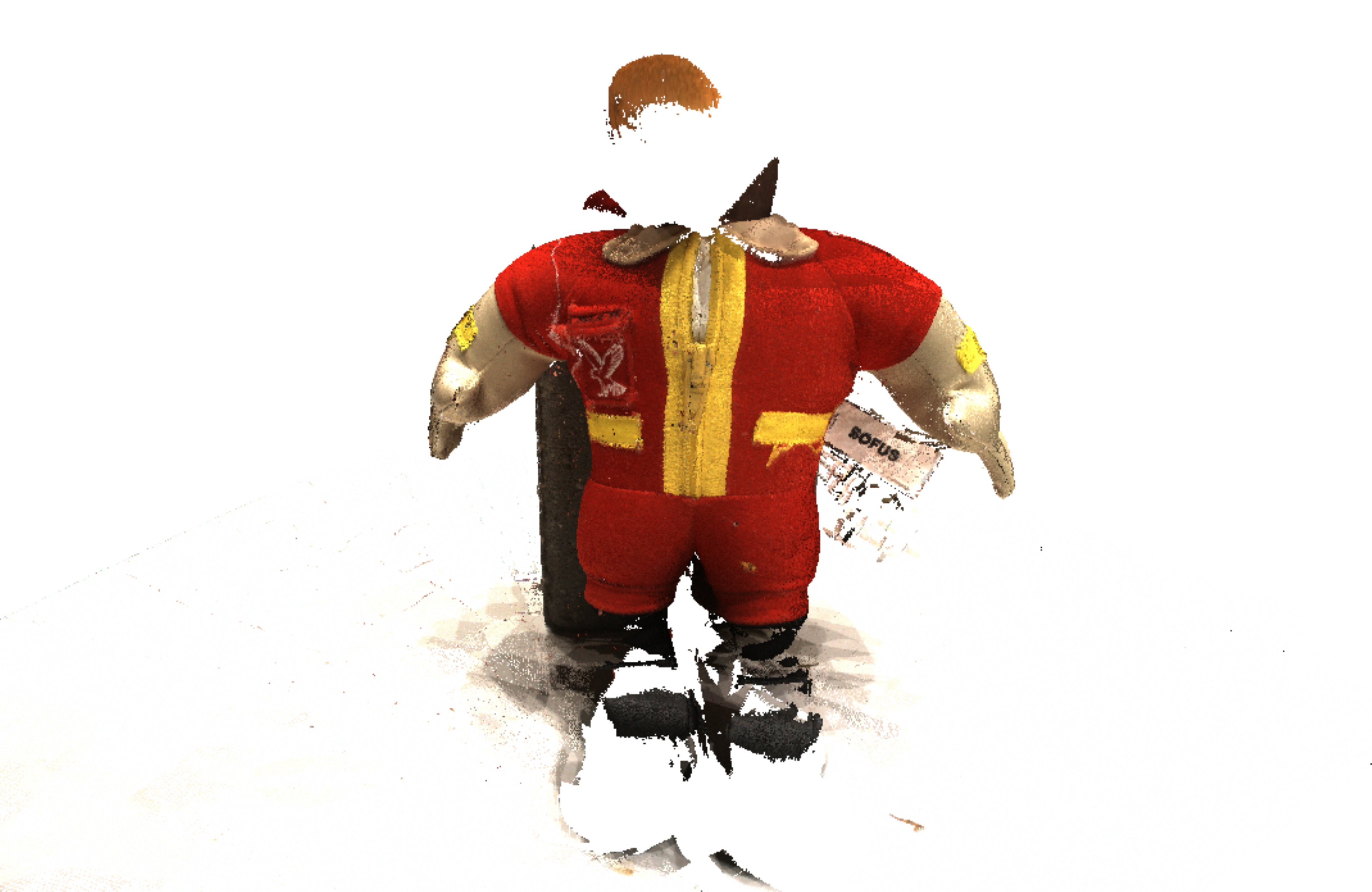}
}
\quad
\subfigure{
\includegraphics[width=0.2\linewidth]{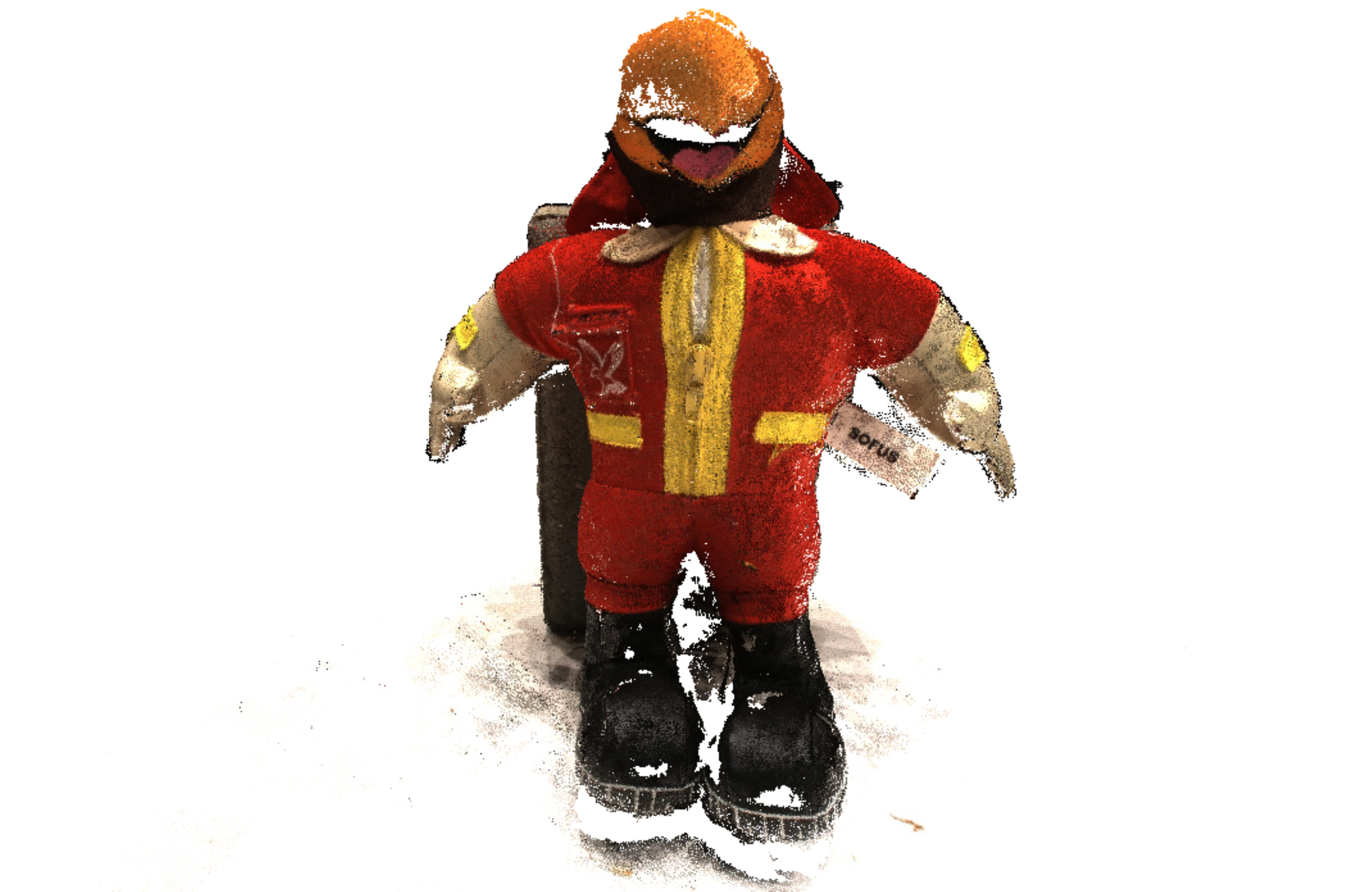}
}
\quad
\subfigure{
\includegraphics[width=0.2\linewidth]{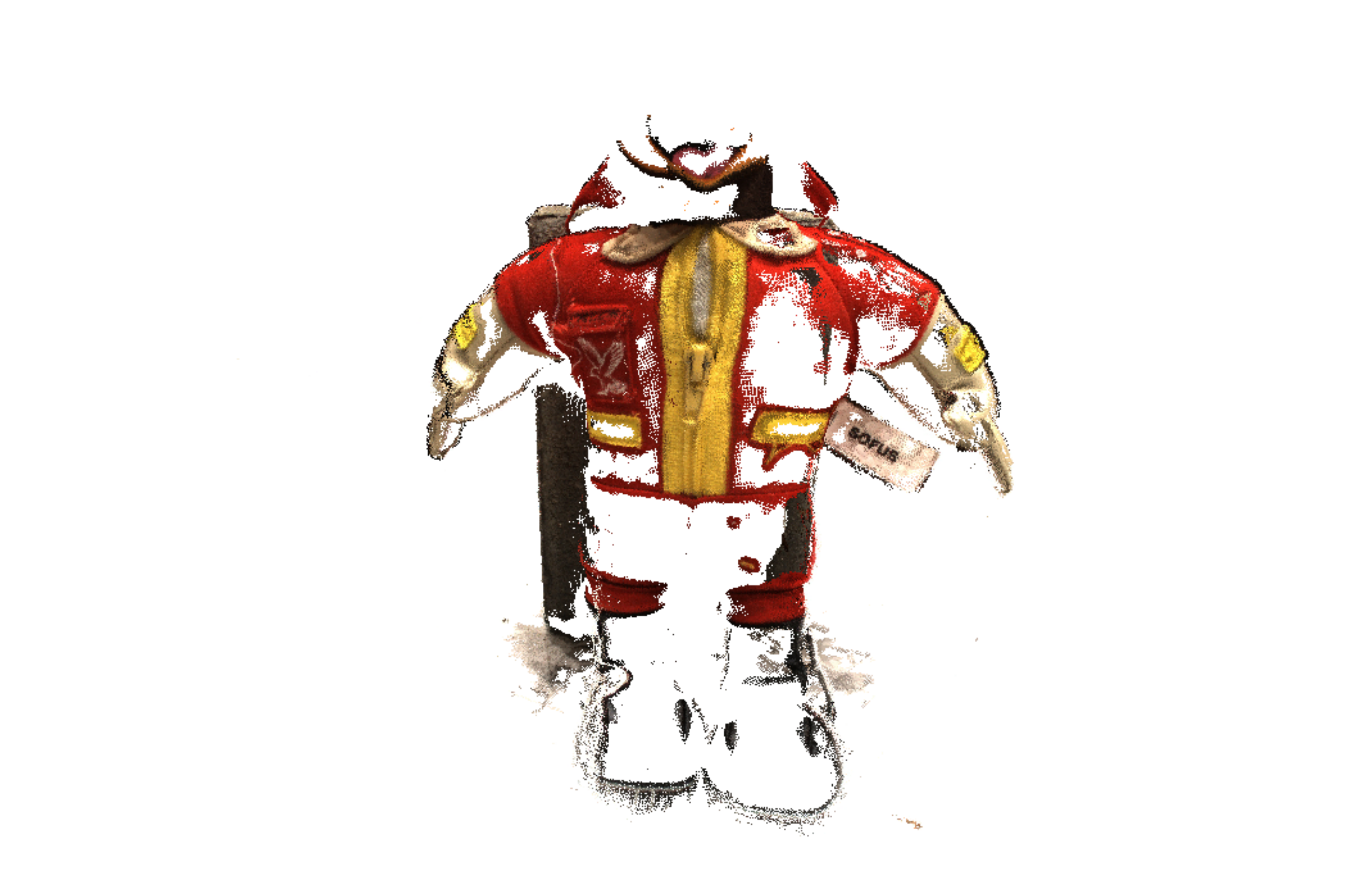}
}
\quad
\subfigure{
\includegraphics[width=0.2\linewidth]{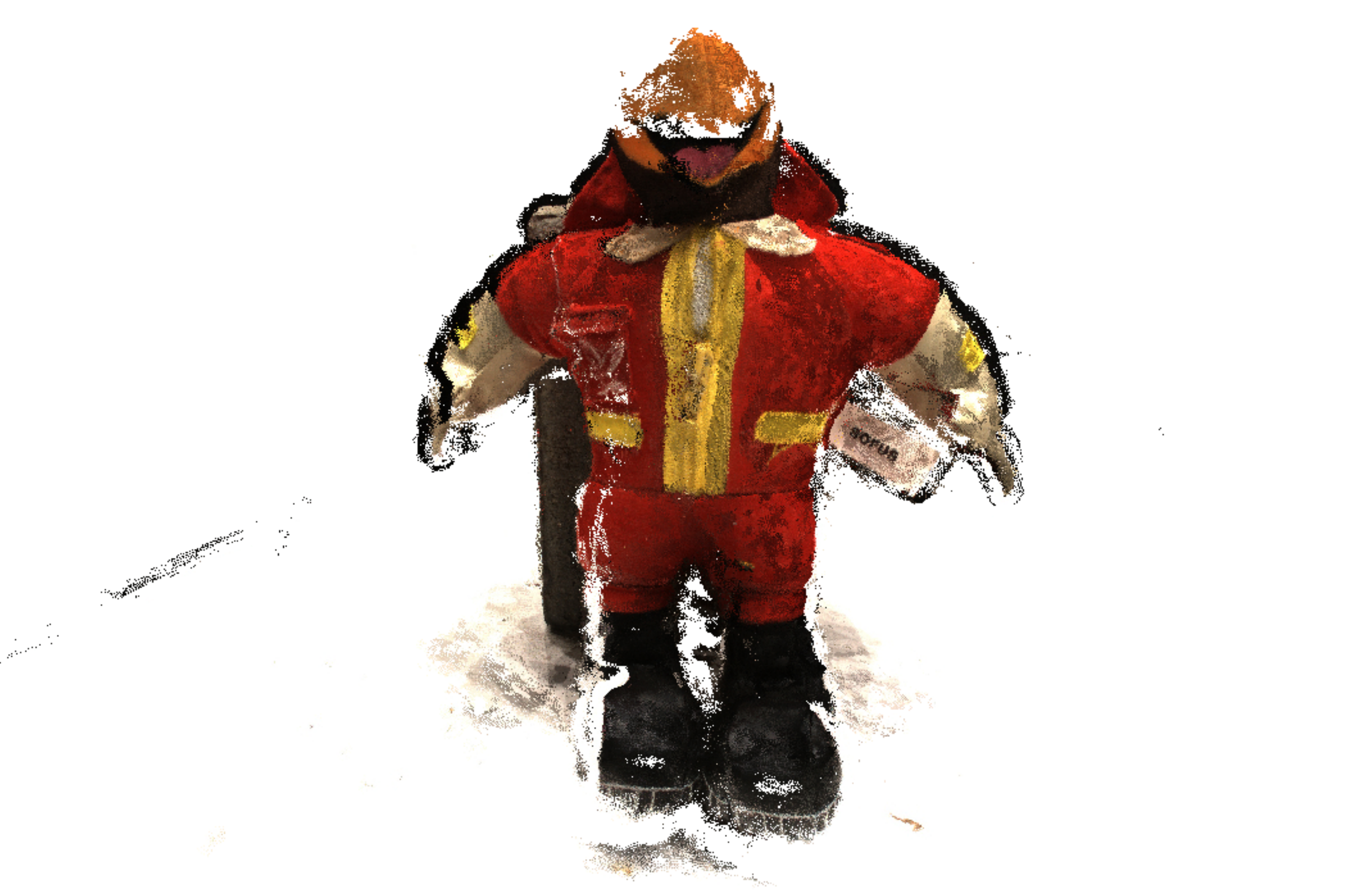}
}
\quad
\subfigure[(a) Ground Truth]{
\includegraphics[width=0.2\linewidth]{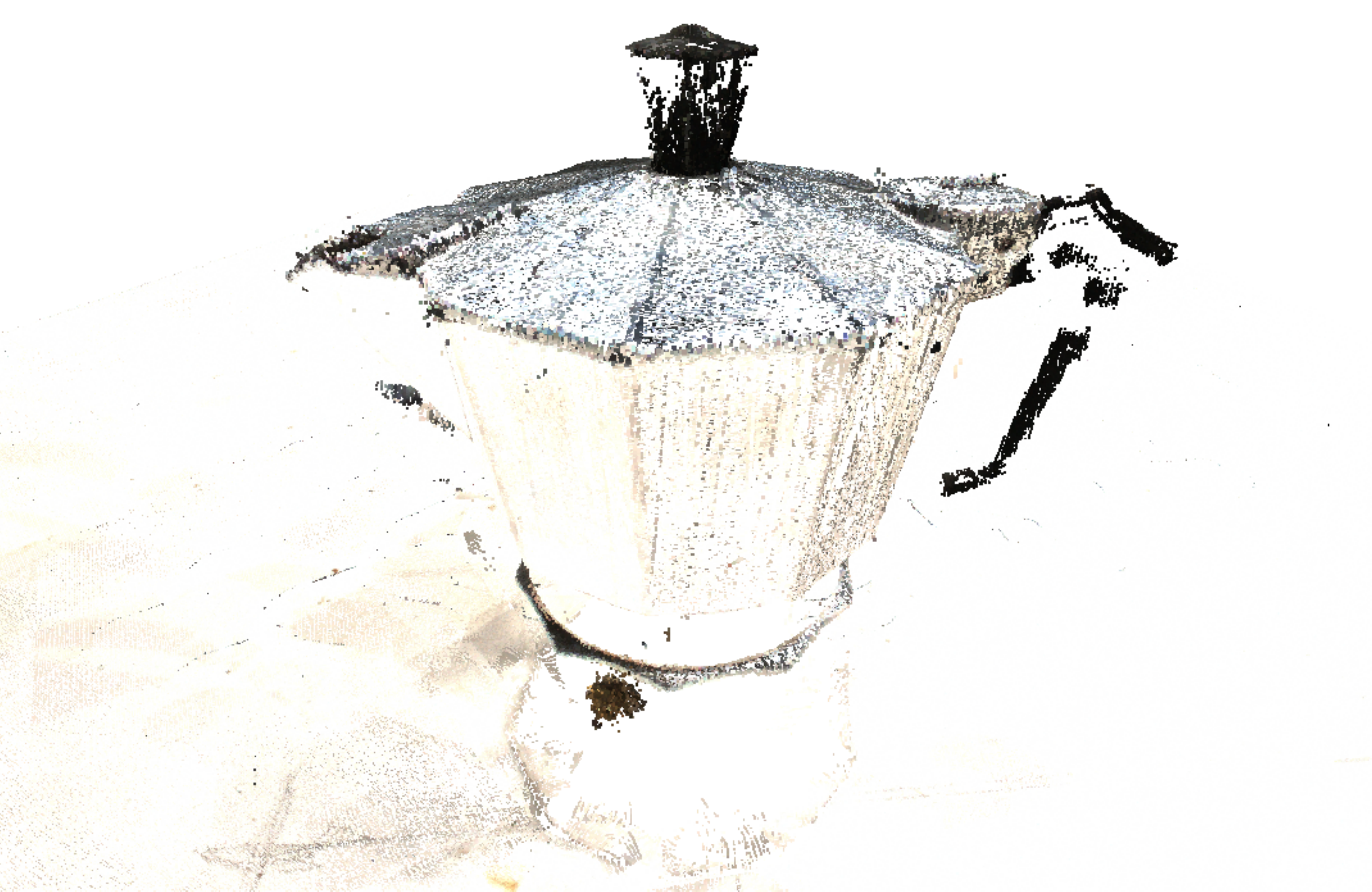}
}
\quad
\subfigure[(b) MVSNet]{
\includegraphics[width=0.2\linewidth]{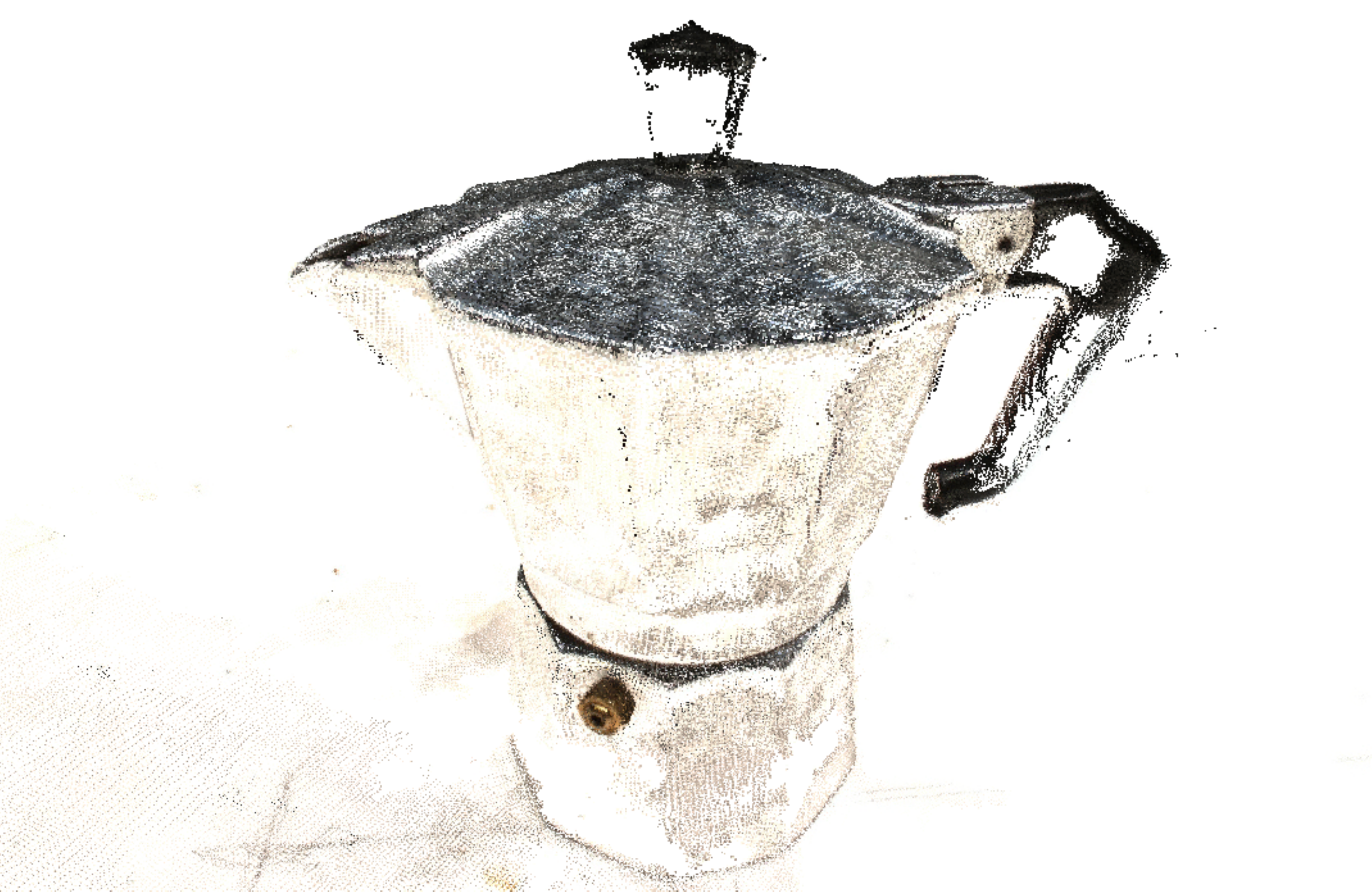}
}
\quad
\subfigure[(c) M$\mathbf{^3}$VSNet w/o Feature-wise Loss]{
\includegraphics[width=0.2\linewidth]{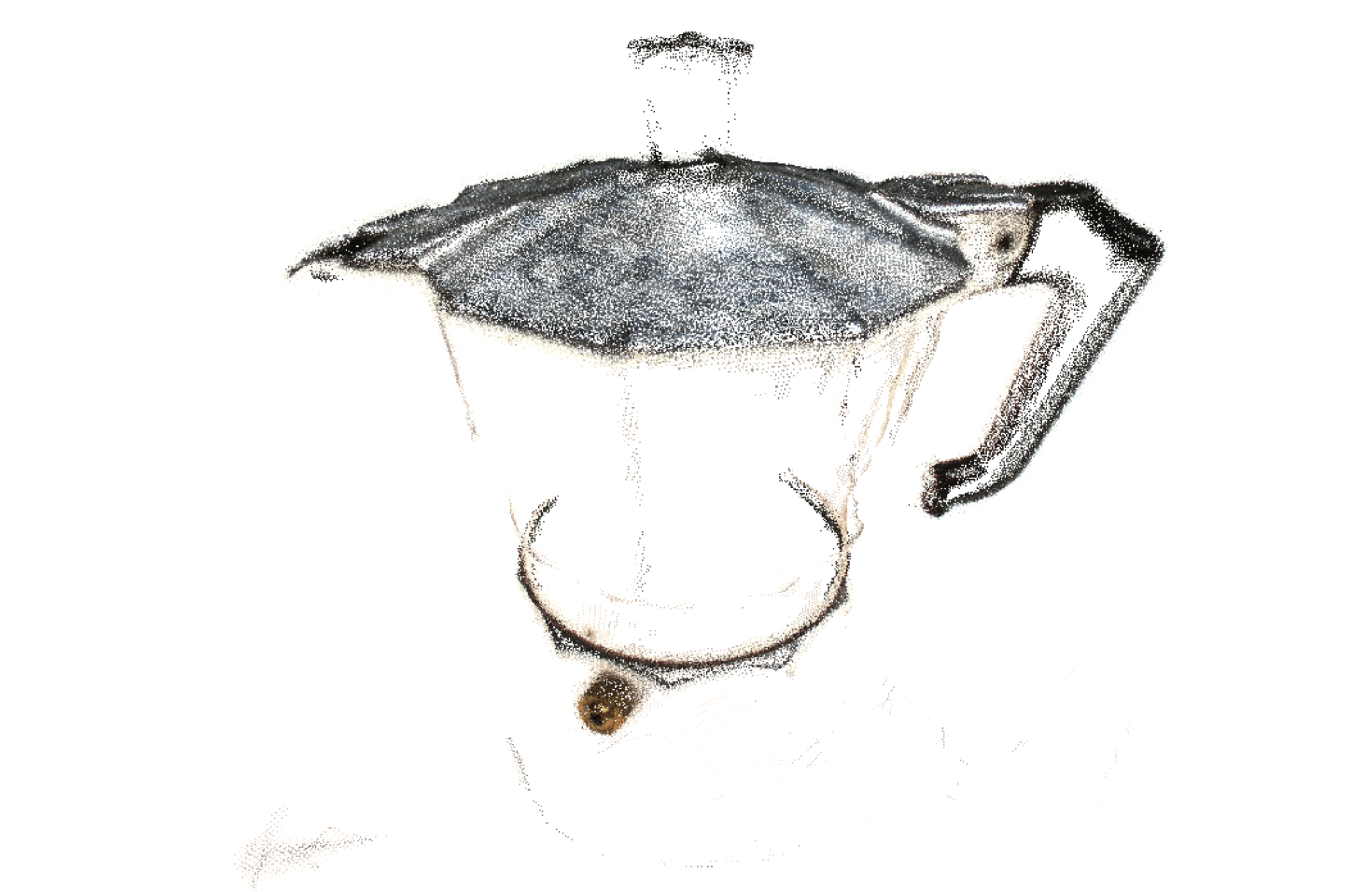}
}
\quad
\subfigure[(d) M$\mathbf{^3}$VSNet]{
\includegraphics[width=0.2\linewidth]{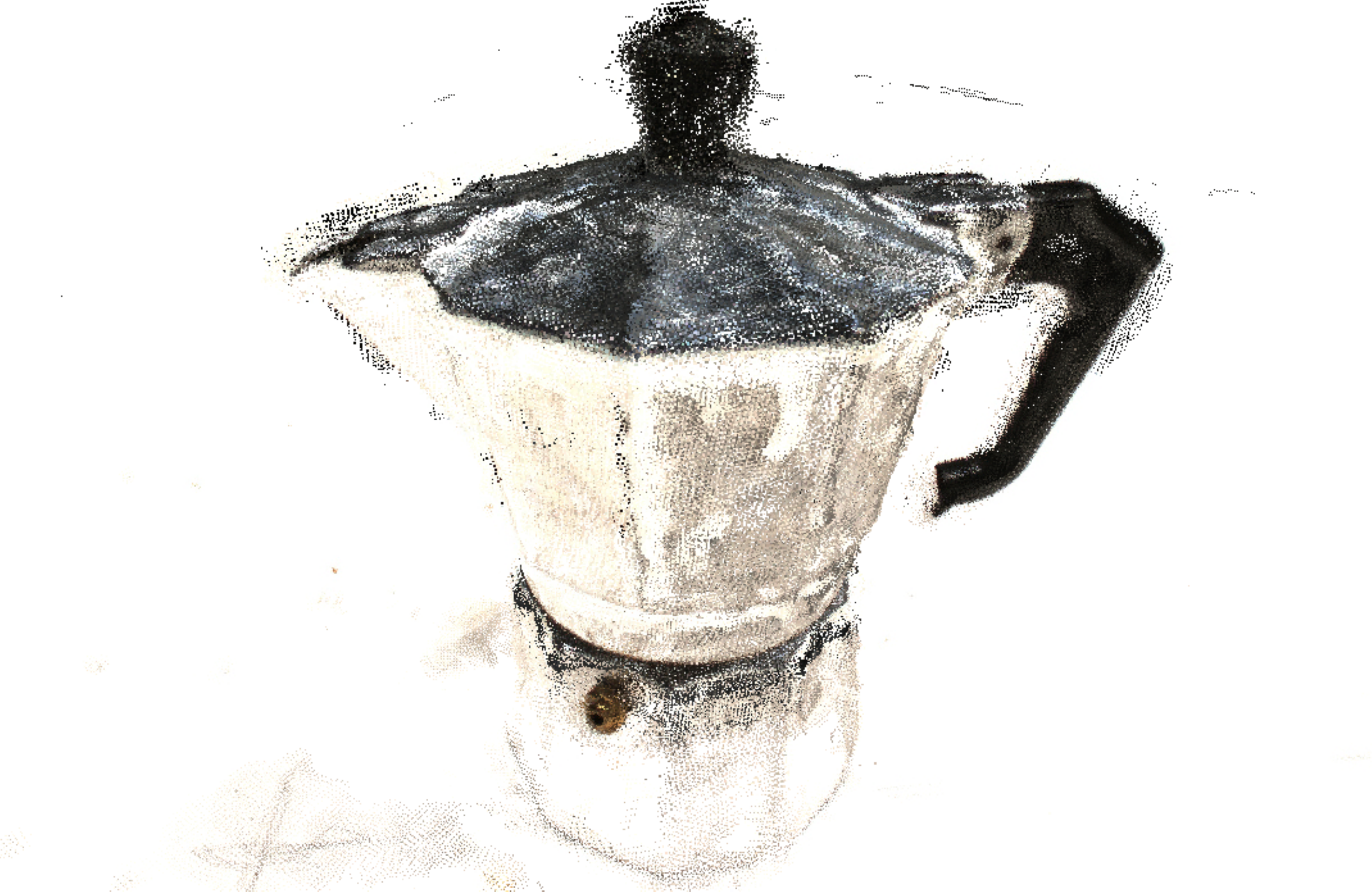}
}
\caption{Qualitative comparison of 3D reconstruction between M$\mathbf{^3}$VSNet and supervised methods on the \textsl{DTU} dataset. From left to right: ground truth, MVSNet \cite{yao2018mvsnet}, M$\mathbf{^3}$VSNet without feature-wise loss and M$\mathbf{^3}$VSNet. Our proposed M$\mathbf{^3}$VSNet establishes the state-of-the-arts unsupervised learning-based method and achieves comparable performance with MVSNet \cite{yao2018mvsnet}.}
\label{fig:comparision}
\end{figure*}

The official metrics \cite{jensen2014large} are used to evaluate M$\mathbf{^3}$VSNet' performance on the \textsl{DTU} dataset. There are three metrics called accuracy, completeness and overall. The overall is the mean value of accuracy and completeness. To prove the effectiveness of M$\mathbf{^3}$VSNet, we compare M$\mathbf{^3}$VSNet with three classic traditional methods such as Furu \cite{Furukawa2007AccurateDA}, Tola \cite{Tola2011EfficientLM} and Colmap \cite{schonberger2016pixelwise}, and with two classic supervised learning-based methods such as SurfaceNet \cite{ji2017surfacenet} and MVSNet \cite{yao2018mvsnet}, and with the other two existed unsupervised learning-based methods such as Unsup\_MVS \cite{khot2019learning} and $\mathrm{MVS^2}$ \cite{dai2019mvs2}.

As shown in the table \ref{tab:resultTable}, our proposed M$\mathbf{^3}$VSNet outperforms the existed two unsupervised learning-based methods \cite{khot2019learning,dai2019mvs2}. M$\mathbf{^3}$VSNet surpasses Unsup\_MVS \cite{khot2019learning} in all metrics and surpasses $\mathrm{MVS^2}$ in accuracy and overall except completeness. Therefore, our proposed M$\mathbf{^3}$VSNet establishes the state-of-the-arts unsupervised learning methods for multi-view stereo reconstruction. Moreover, M$\mathbf{^3}$VSNet surpasses the supervised learning-based MVSNet \cite{yao2018mvsnet} with the same setting depth hypothesis $D=192$ in terms of the overall performance of point cloud reconstruction. Compared with traditional MVS methods \cite{frahm2010building,Tola2011EfficientLM,schonberger2016pixelwise}, our proposed M$\mathbf{^3}$VSNet achieves significant improvement on the completeness of point cloud reconstruction and outperforms Furu \cite{Furukawa2007AccurateDA} and Tola \cite{Tola2011EfficientLM} on the overall quality except Colmap \cite{schonberger2016pixelwise} but with high efficiency. For more detailed information in point cloud reconstruction, figure \ref{fig:comparision} illustrates the qualitative comparison. The reconstruction by M$\mathbf{^3}$VSNet has more complete texture details than that without feature-wise loss. With the aid of multi-metric, M$\mathbf{^3}$VSNet is more robust so that it recovers more textureless or texture repeat areas and achieves comparable visual performance with original MVSNet \cite{yao2018mvsnet}.

\begin{table}[t]
\caption{Quantitative results on the \textsl{DTU}’s evaluation set. Three classical MVS methods, two supervised learning-based MVS methods and three unsupervised methods using the distance metric (lower is better) are listed.}\label{tab:resultTable}
\begin{center}
\begin{tabular}{cccc}
\toprule
\multicolumn{1}{c}{Method}&\multicolumn{3}{c}{Mean Distance (mm)}\\
 & Acc. & Comp. & overall.\\
\midrule
Furu \cite{Furukawa2007AccurateDA}& 0.612 & 0.939 & 0.775 \\
Tola \cite{Tola2011EfficientLM} & $\textbf{0.343}$ & 1.190 & 0.766\\
Colmap \cite{schonberger2016pixelwise} & 0.400 & $\textbf{0.664}$ & $\textbf{0.532}$\\
\midrule
SurfaceNet \cite{ji2017surfacenet} & 0.450 & 1.043 & 0.746\\
MVSNet(D=192) & $\textbf{0.444}$ & $\textbf{0.741}$ & $\textbf{0.592}$\\
\midrule
Unsup\_MVS \cite{khot2019learning} & 0.881 & 1.073 & 0.977\\
$\mathrm{MVS^2}$ \cite{dai2019mvs2} & 0.760 & $\textbf{0.515}$ & 0.637\\
$\textbf{M$\mathbf{^3}$VSNet(D=192)}$ & $\textbf{0.636}$ & 0.531 & $\textbf{0.583}$\\
\bottomrule
\end{tabular}
\end{center}
\end{table}

\subsection{Comparison With Unsupervised Methods}

M$\mathbf{^3}$VSNet establishes the state-of-the-art unsupervised learning-based MVS network by outperforming the other two existing unsupervised MVS networks \cite{khot2019learning,dai2019mvs2}. One is unsup\_mvs \cite{khot2019learning}, which is almost the first try in this direction but with poor performance where the overall mean distance is 0.977. The other one is $\mathrm{MVS^2}$ \cite{dai2019mvs2}. Although $\mathrm{MVS^2}$ can get a little bit better completeness than M$\mathbf{^3}$VSNet and can reach to 0.637 in the overall mean distance, it consumes more GPU memory due to three cost volumes and regularization needed to be constructed, which is unaffordable for a single NVIDIA RTX 2080Ti used in M$\mathbf{^3}$VSNet. As a result, our proposed unsupervised method achieves the best performance on the overall quality of point cloud reconstruction with high efficiency where the accuracy of point cloud is significantly improved.

\subsection{Ablation Studies}
\label{sec:ablation}
The section begins to analyze the effect of different modules proposed in M$\mathbf{^3}$VSNet. There are mainly three contrast experiments carried out. We will explore the effect of pyramid feature aggregation, normal-depth consistency and multi-metric loss.

\paragraph{\textbf{Pyramid feature aggregation}}

The module can catch more contextual information from low-level to high-level representations. We use the feature pyramid aggregation to output the 1/4 feature. By pyramid feature aggregation, the matching correspondences will be guaranteed to a large extent. As shown in table \ref{tab:pyramidfeatureTable}, this module will improve the metric of accuracy and completeness in mean distance. Further, pyramid feature aggregation improves 2\% in overall.

\begin{table}[t]
\caption{Comparison of the performance in pyramid feature aggregation using the distance metric (lower is better).}
\label{tab:pyramidfeatureTable}
\begin{center}
\begin{tabular}{cccc}
\toprule
\multicolumn{1}{c}{Method}&\multicolumn{3}{c}{Mean Distance (mm)}\\
 & Acc. & Comp. & overall\\
\midrule
without pyramid feature aggregation & 0.638 & 0.554 & 0.596 \\
$\textbf{with pyramid feature aggregation}$ & $\textbf{0.636}$ & $\textbf{0.531}$ & $\textbf{0.583}$\\
\bottomrule
\end{tabular}
\end{center}
\end{table}

\paragraph{\textbf{Normal-depth consistency}}

Based on the orthogonality between local surface tangent and normal, normal-depth consistency is introduced to regularize the depth in 3D space. Absolute depth error is used to evaluate the quality of estimated depth. Here we use the percentage of depth error within 2mm, 4mm, and 8mm compared with ground-truth depth maps (Higher is better). As shown in table \ref{tab:NormaldepthTable}, the performance with the aid of normal-depth consistency surpasses that without normal-depth consistency in all metrics.

\begin{table}[t]
\caption{Comparison of the performance in normal-depth consistency using the depth error (higher is better).}
\label{tab:NormaldepthTable}
\begin{center}
\begin{tabular}{cccc}
\toprule
%\multicolumn{1}{c}{Method}&\multicolumn{3}{c}{Depth Error %(mm)}\\
Depth Error (mm) & $\%<2$ & $\%<4$ & $\%<8$\\
\midrule
without normal-depth consistency& 58.8 & 74.8 & 83.8 \\
$\textbf{with normal-depth consistency}$ & $\textbf{60.3}$ & $\textbf{76.9}$ & $\textbf{85.7}$\\
\bottomrule
\end{tabular}
\end{center}
\end{table}

\begin{figure}[t]
\begin{center}
\includegraphics[width=1.0\linewidth]{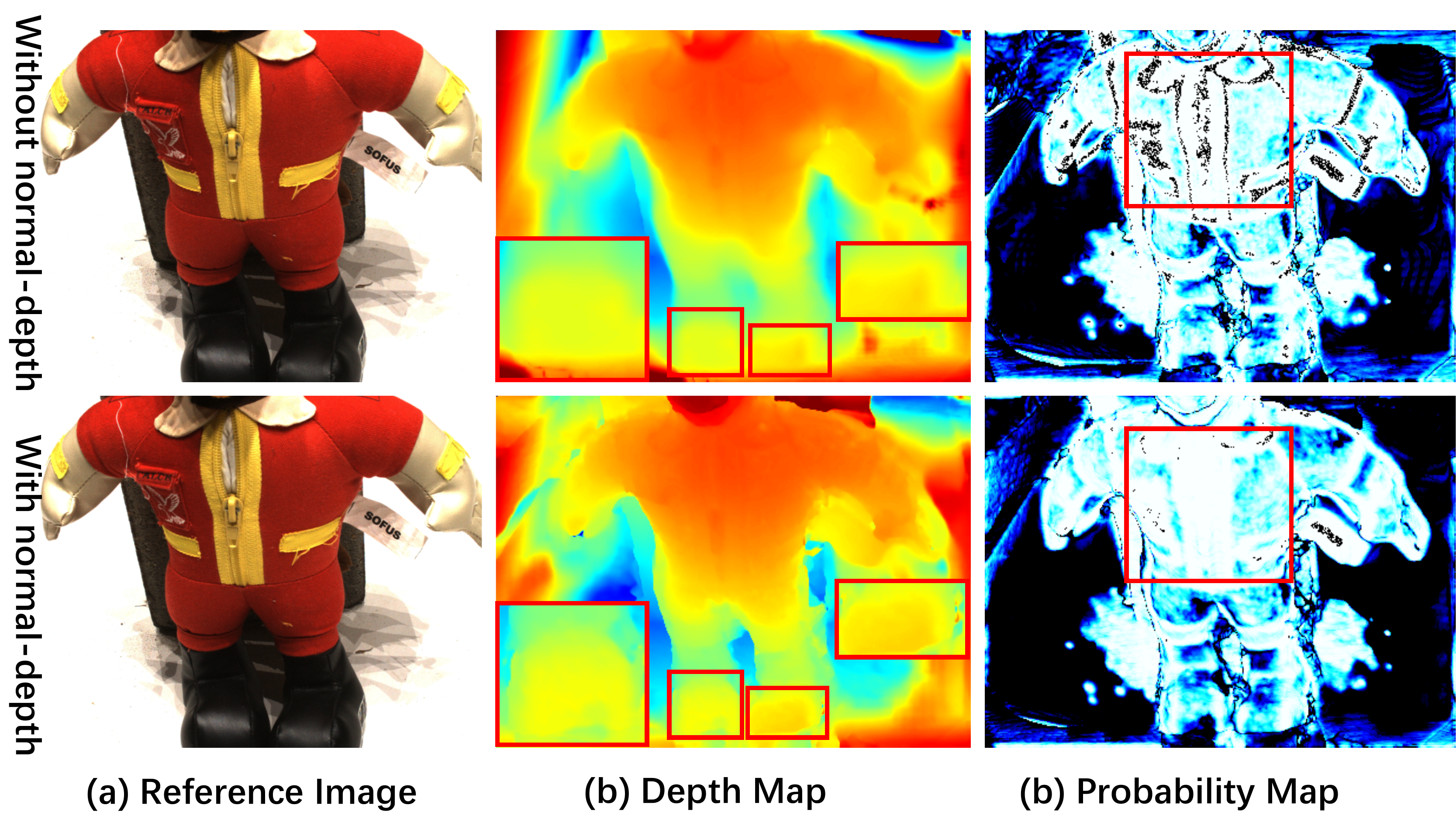}
\end{center}
   \caption{Qualitative comparison of normal-depth consistency in depth maps and probability maps (White is 100\%, black is 0\%). The performance of M$\mathbf{^3}$VSNet will be more robust and more accurate even in non-ideal environments.}
\label{fig:normaldepth}
\end{figure}

Figure \ref{fig:normaldepth} demonstrates the comparison with and without normal-depth consistency. For the same reference image, the depth map with normal-depth consistency is more accurate than that without normal-depth consistency. The depth of foot (the part framed in red) in the first row is more ambiguous than that in the second row along the edge and the plane. What's more, the foot part is textureless with reflective effect. Therefore, the robustness of performance with normal-depth consistency will be guaranteed even in non-ideal environments. Normal-depth consistency will make the estimated depth more precise and reasonable in 3D space. Furthermore, in probability maps, the probability of small local areas such as the collar and the zipper (the part framed in red) will be high with the aid of normal-depth consistency. The module also improves the probability of correct matching correspondences in some different planes. Figure \ref{fig:normaldepth} and table \ref{tab:NormaldepthTable} can prove the significant benefits of normal-depth consistency for accuracy and continuity.

\paragraph{\textbf{Multi-metric loss}}

Multi-metric loss contains pixel-wise loss and feature-wise loss, which learns the inherent constraints from different perspectives of matching correspondences. The try to feature-wise loss is effective in previous related works \cite{johnson2016perceptual}\cite{wang2018depth}\cite{benzhang2018unsupervised}. We have compared the performance of different combinations of pixel-wise loss and feature-wise loss. What's more, how to select the multi-scale features is also taken into consideration. 

In the ablation study of loss terms in pixel-wise loss, the absolute depth error is used to evaluate the performance of different loss terms. As shown in table \ref{tab:loss_term}, B is the baseline with the only photometric loss in pixel level. G represents the gradient consistency loss. As demonstrated in section \ref{sec:multi-metric}, SSIM is represented as $L_{SSIM}$ and smooth is represented as $L_{smooth}$. The terms of G and SSIM improve the results slightly and the term of Smooth contributes a lot with effective improvement. In general, it's apparent that the proposed each loss will improve the performance of M$\mathbf{^3}$VSNet. 

\begin{table}[t]
\caption{Comparison of the performance in the different loss terms using the percentage of depth error (higher is better).}
\label{tab:loss_term}
\begin{center}
\begin{tabular}{cccc}
\toprule
%\multicolumn{1}{c}{Method}&\multicolumn{3}{c}{Depth Error %(mm)}\\
Depth Error (mm) & $\%<2$ & $\%<4$ & $\%<8$\\
\midrule
B & 22.1 & 36.5 & 50.8 \\
B+G & 25.2 & 40.7 & 55.3 \\
B+G+SSIM & 27.5 & 44.2 & 58.8 \\ 
B+G+SSIM+Smooth & 57.5 & 75.2 & 85.4 \\
$\textbf{Multi-metric loss}$ & $\textbf{60.3}$ & $\textbf{76.9}$ & $\textbf{85.7}$\\
\bottomrule
\end{tabular}
\end{center}
\end{table}

In the ablation study of loss terms in feature-wise loss, as illustrated in table \ref{tab:featurewiseTable}, the overall of only pixel-wise loss is relatively higher (lower is better). Besides, the different combinations of feature-wise losses make it an impressive improvement. We do some ablation studies on the different combinations of features from pre-trained VGG16. Adding the 1/16 feature improves the accuracy but deteriorate the completeness. By comparison, the combination of 1/2, 1/4, 1/8 features achieves the best result.

\begin{table}[t]
\caption{Comparison of the performance in different loss (lower is better). The scale of 1/2 represents that the feature (corresponding to layer 8) extracted from the pre-trained VGG16 networks is half the size of the original reference image. The scales of 1/4, 1/8, 1/16 correspond to layer 15, 22, 29.}
\label{tab:featurewiseTable}
\begin{center}
\begin{tabular}{cccc}
\toprule
\multicolumn{1}{c}{Method}&\multicolumn{3}{c}{Mean Distance (mm)}\\
 & Acc. & Comp. & overall\\
\midrule
only pixel-wise & 0.832 & 0.924 & 0.878 \\
pixel-wise+1/4 feature & 0.646 & 0.591 & 0.618\\
$\textbf{pixel-wise+1/2,1/4,1/8 feature}$ & $\textbf{0.636}$ & $\textbf{0.531}$ & $\textbf{0.583}$\\
pixel-wise+1/2,1/4,1/8,1/16 feature & 0.566 & 0.653 & 0.609\\
\bottomrule
\end{tabular}
\end{center}
\end{table}

\subsection{Generalization Ability on \textsl{Tanks \& Temples}}

\begin{table*}[t]
\caption{Quantitative comparison of point cloud reconstruction on the \textsl{Tanks and Temples} benchmark (higher is better). M$\mathbf{^3}$VSNet surpasses other unsupervised methods by the mean score in the leaderboard of intermediate \textsl{T\&T} \cite{knapitsch2017tanks}.}
\label{tab:TTtable}
\begin{center}
\begin{tabular}{cccccccccc}
\toprule
%\multicolumn{1}{c}{Method}&\multicolumn{3}{c}{Depth Error %(mm)}\\
Method & Mean & Family & Francis & Horse & Lightouse & M60 & Panther & Playground & Train\\
\midrule
$\textbf{M$\mathbf{^3}$VSNet}$ & $\textbf{37.67}$ & $\textbf{47.74}$ & $\textbf{24.38}$ & 18.74	& 44.42 & 43.45	& 44.95	& $\textbf{47.39}$ & $\textbf{30.31}$\\
$\mathrm{MVS^2}$ & 37.21 & 47.74 & 21.55 & $\textbf{19.50}$ & $\textbf{44.54}$ & $\textbf{44.86}$	& $\textbf{46.32}$ & 43.48 & 29.72\\
\bottomrule
\end{tabular}
\end{center}
\end{table*}

\begin{figure*}[t]
\centering
\subfigure[Family]{
\includegraphics[width=0.22\linewidth]{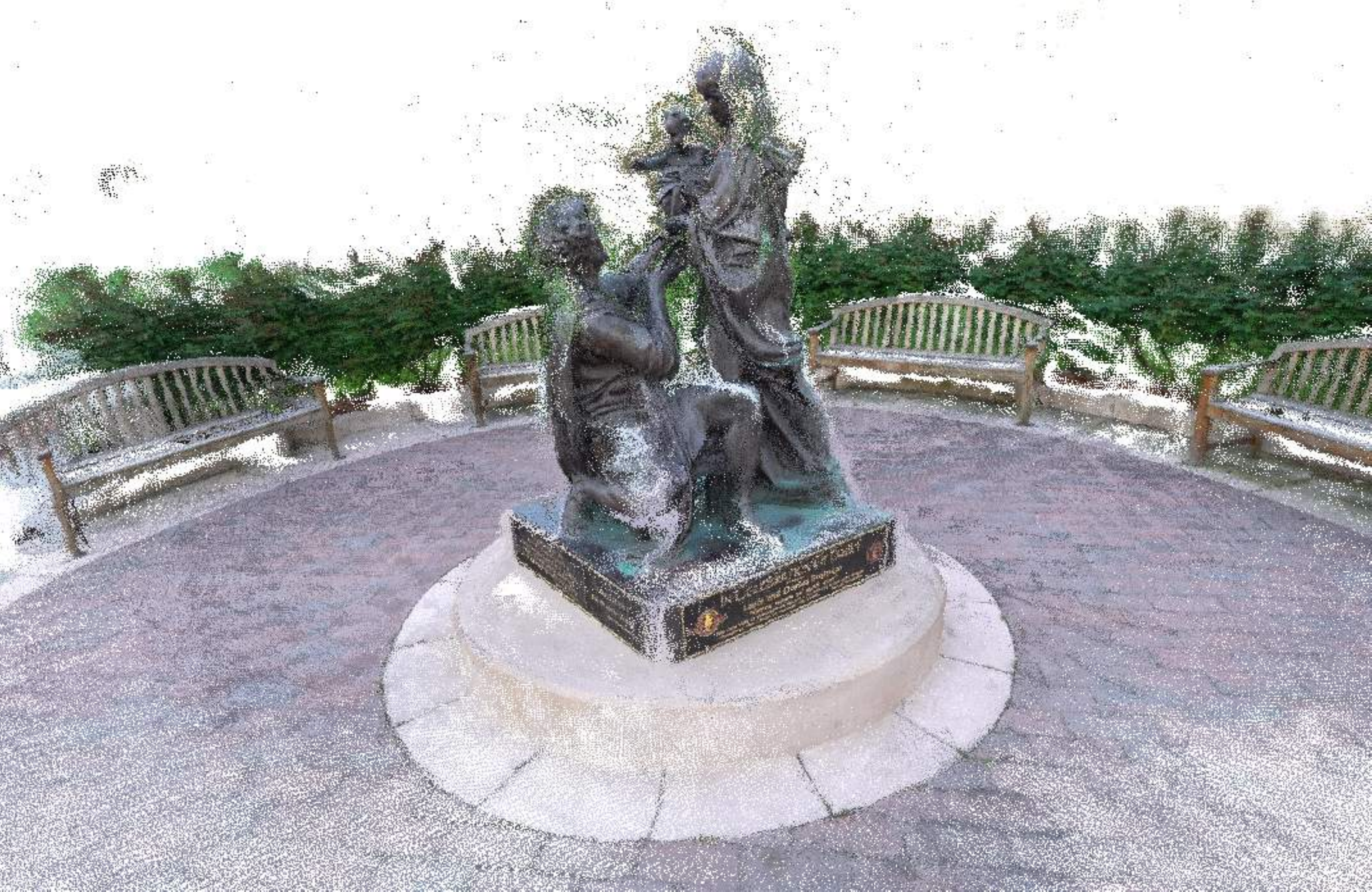}
%\caption{fig1}
}
\quad
\subfigure[Francis]{
\includegraphics[width=0.22\linewidth]{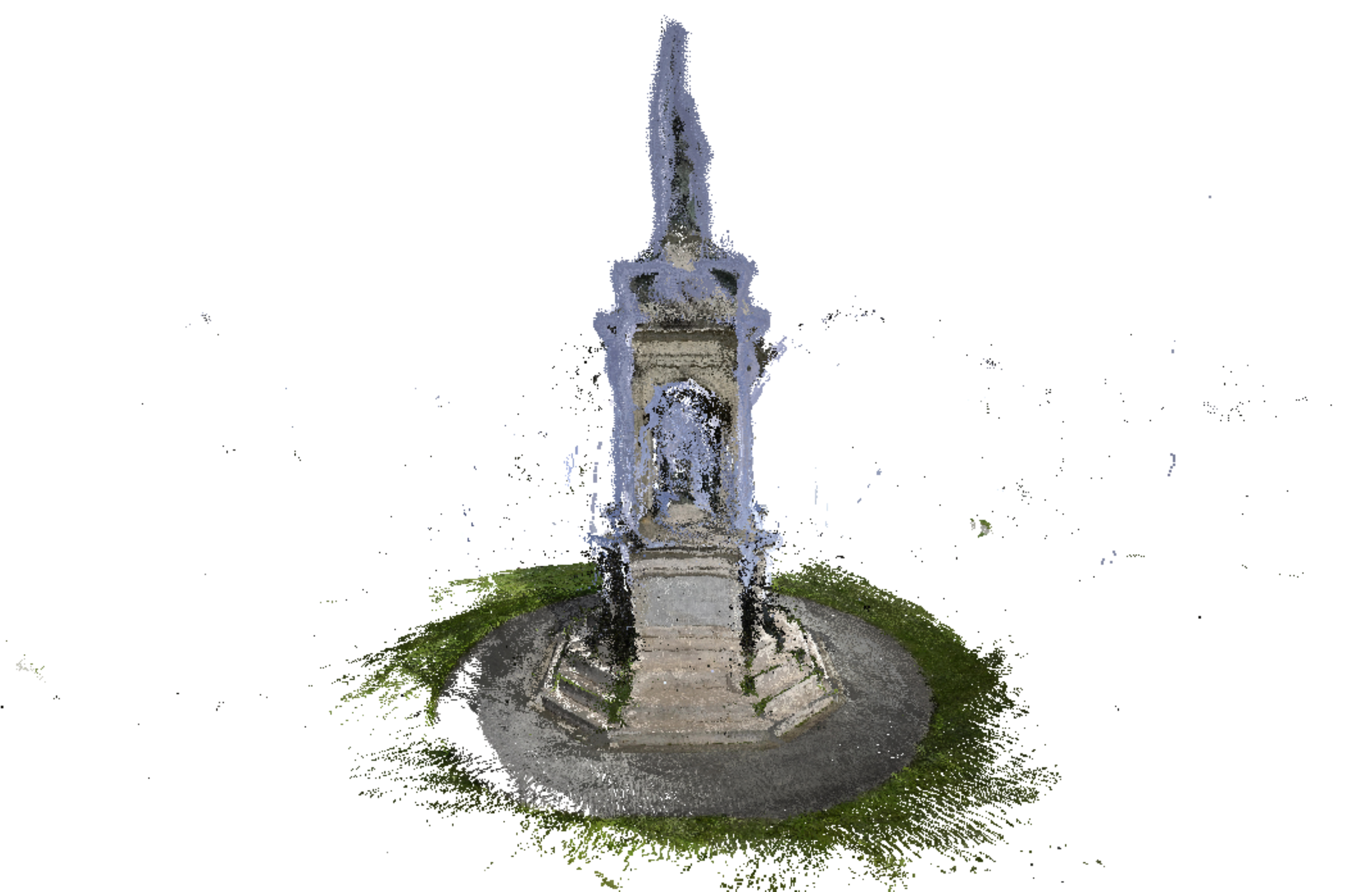}
}
\quad
\subfigure[Horse]{
\includegraphics[width=0.22\linewidth]{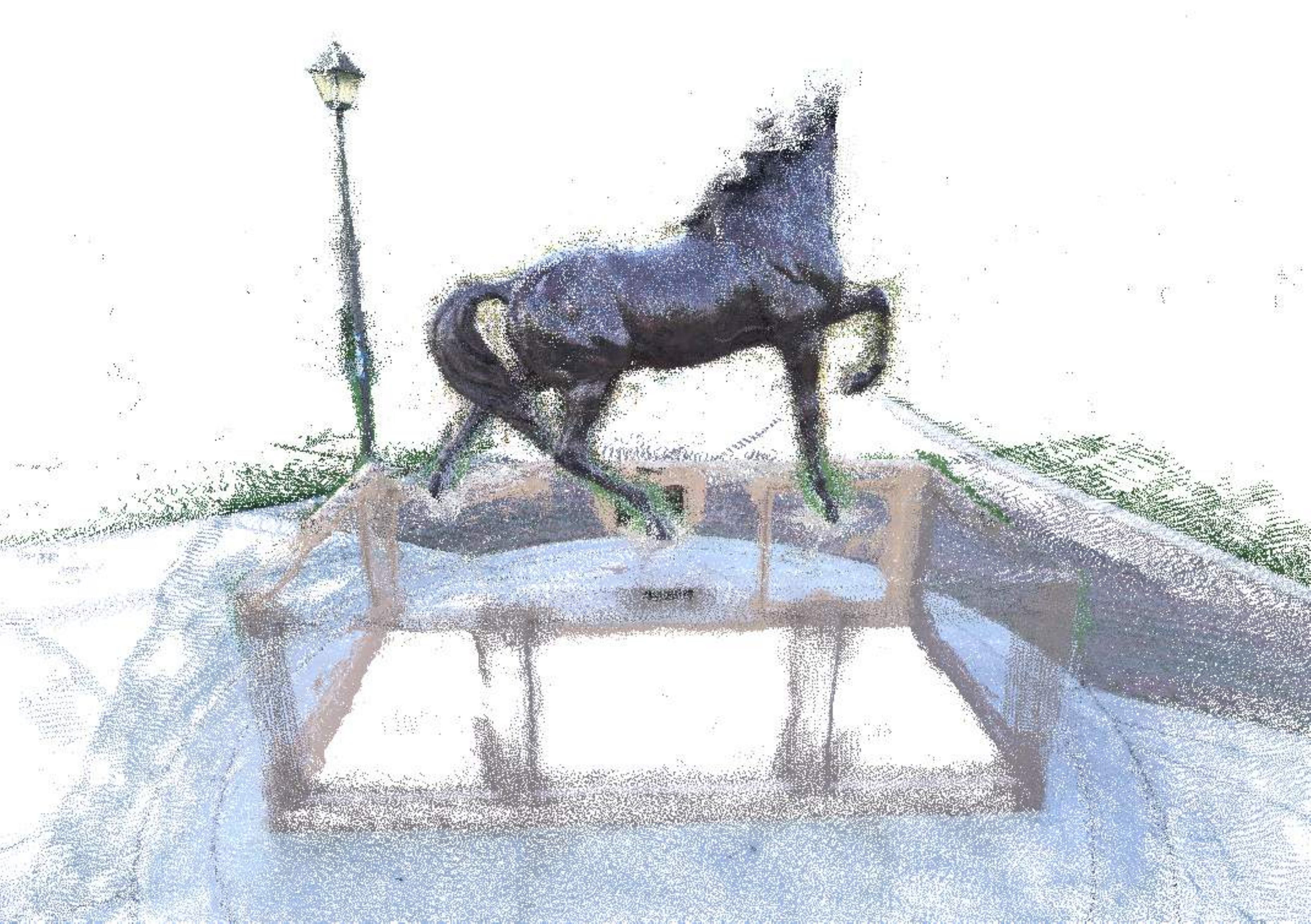}
}
\quad
\subfigure[M60]{
\includegraphics[width=0.22\linewidth]{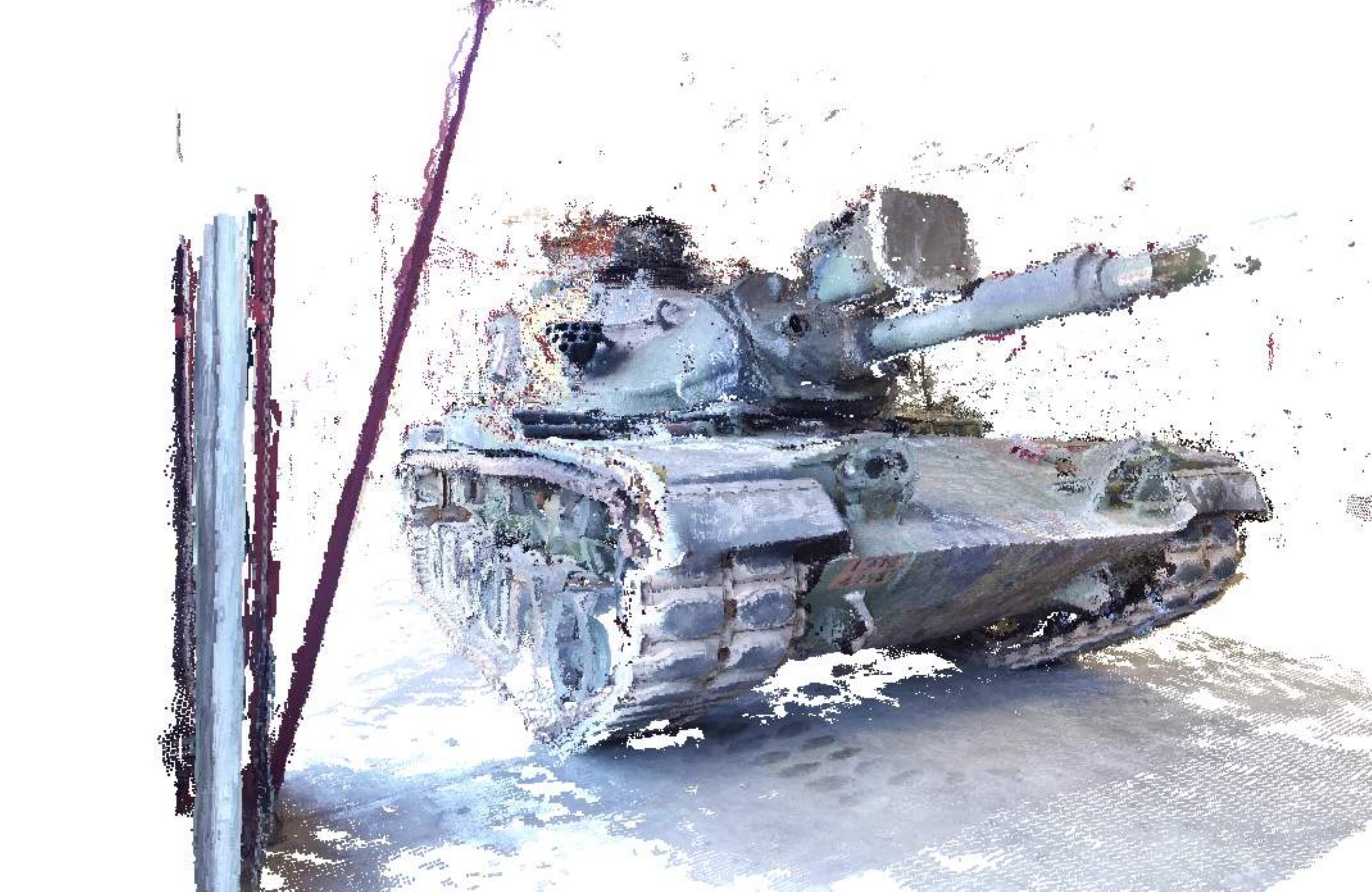}
}
\quad
\subfigure[Panther]{
\includegraphics[width=0.22\linewidth]{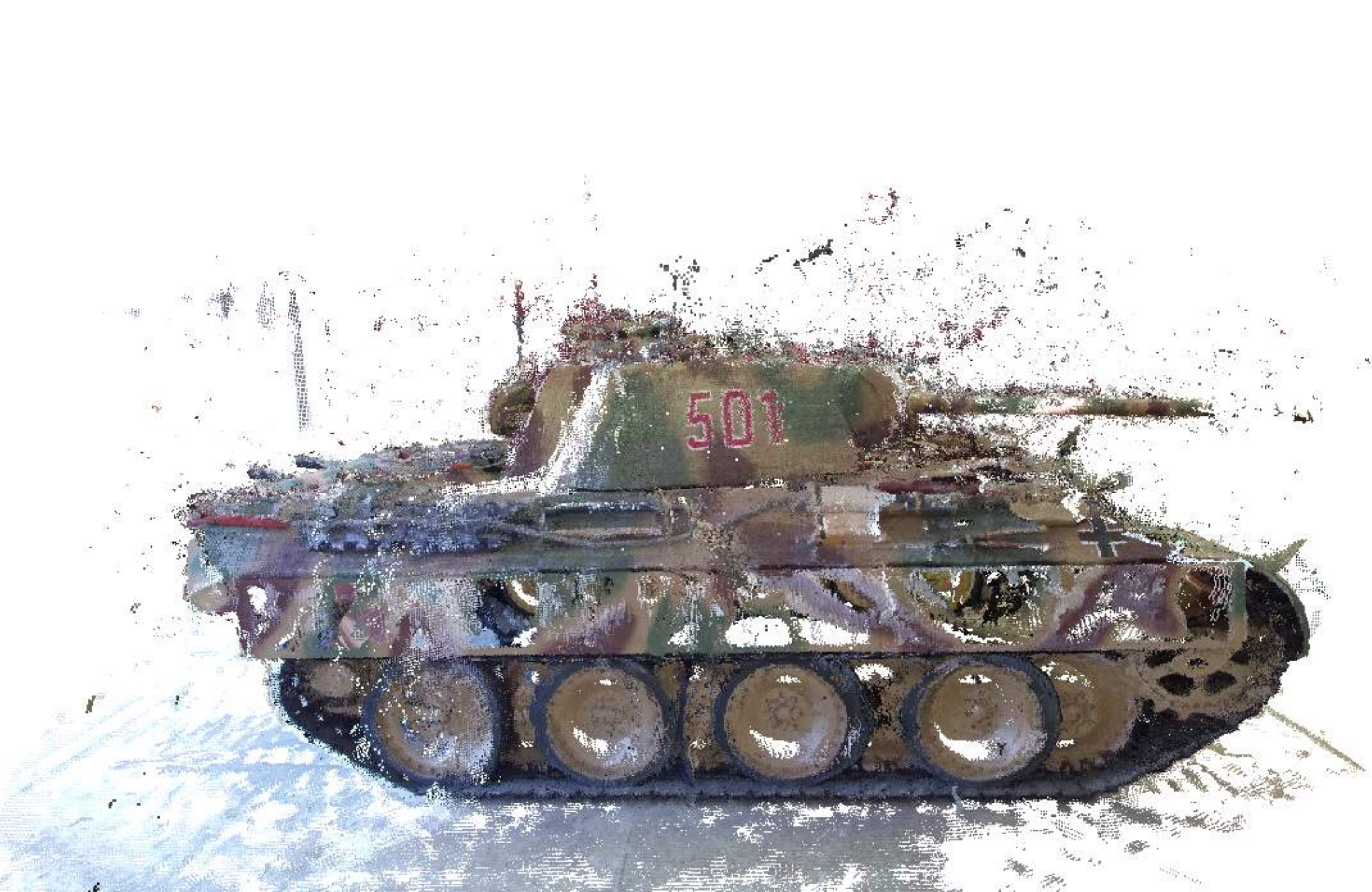}
}
\quad
\subfigure[Playground]{
\includegraphics[width=0.22\linewidth]{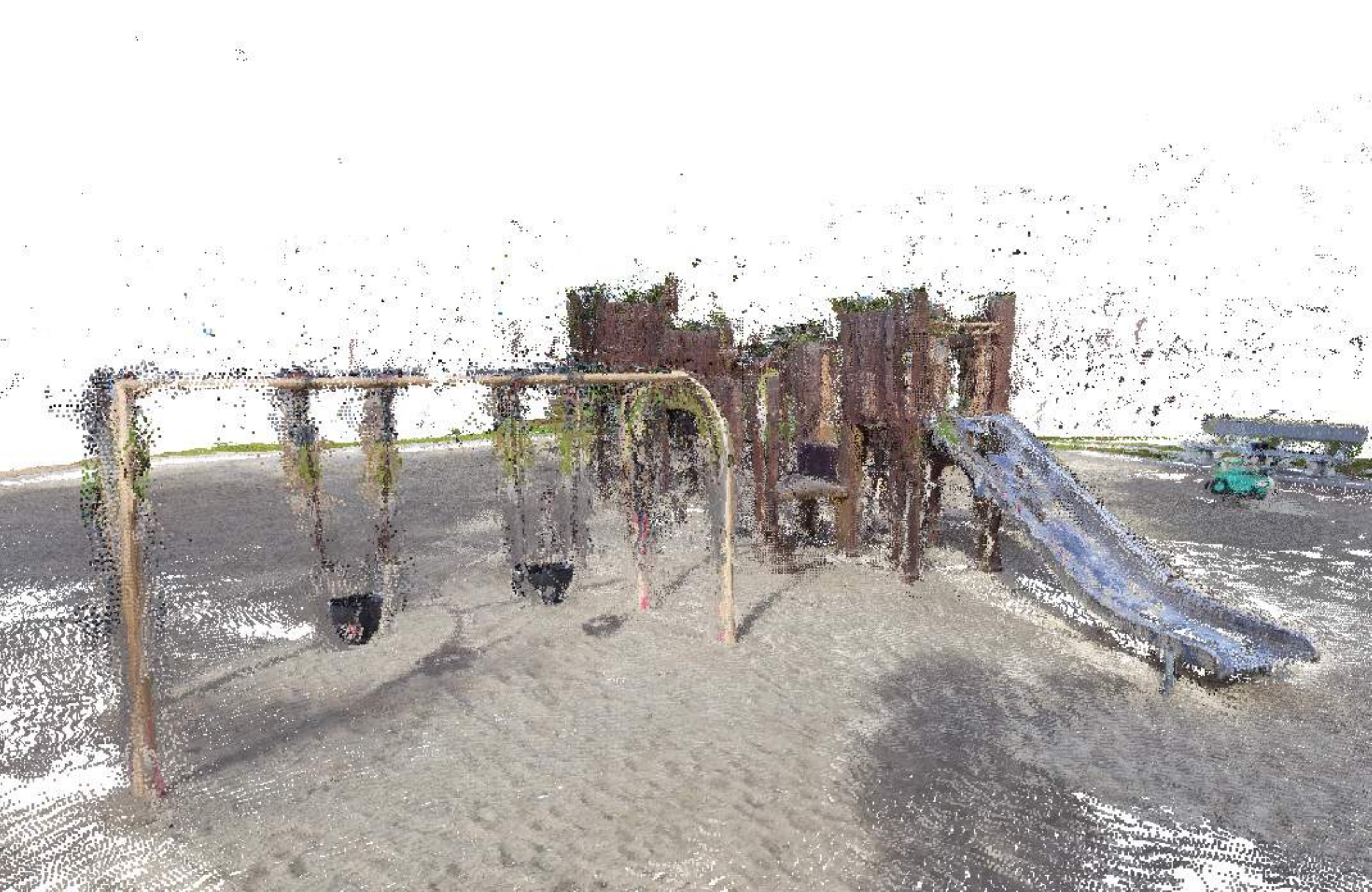}
}
\quad
\subfigure[Train]{
\includegraphics[width=0.22\linewidth]{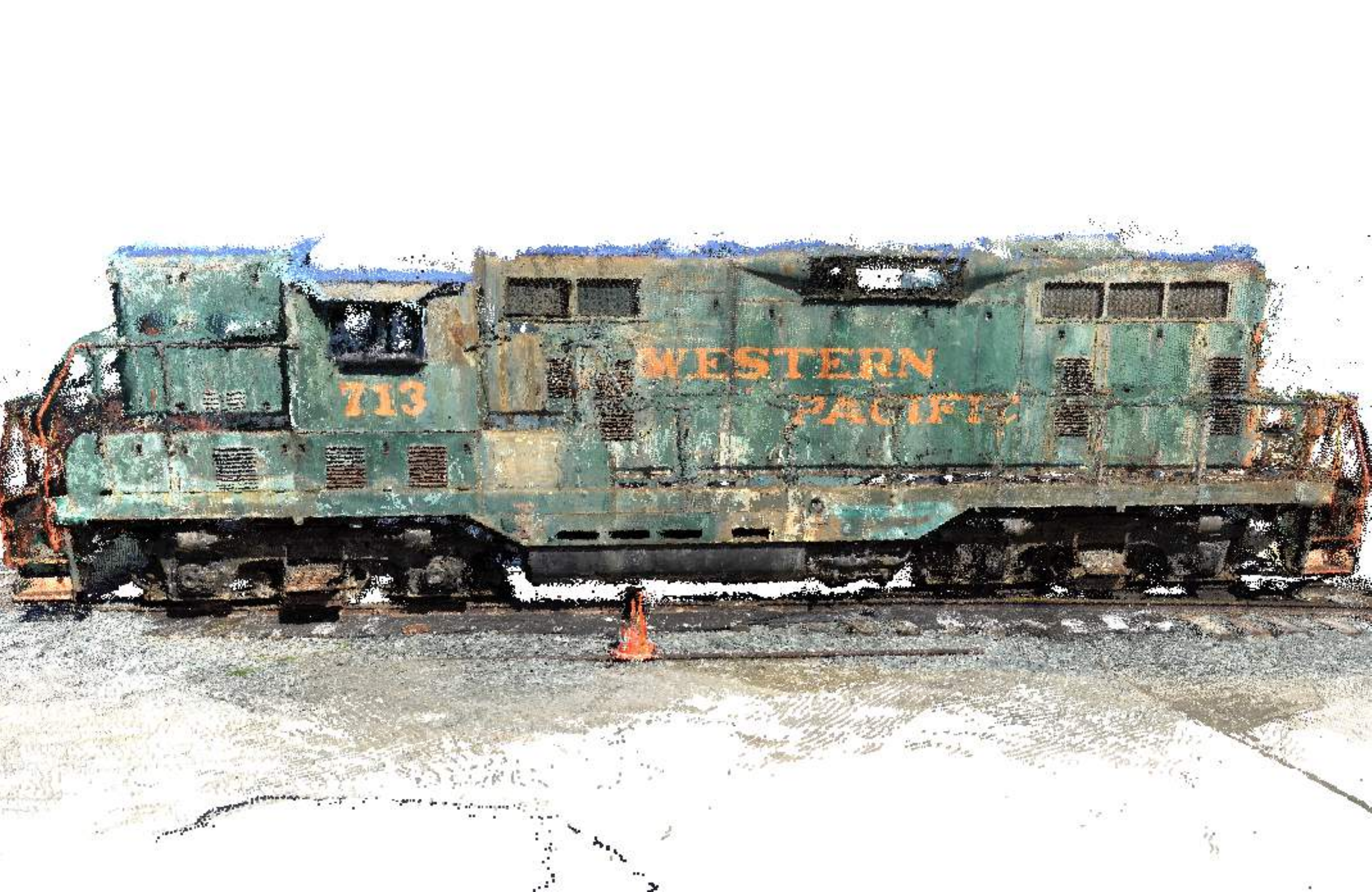}
}
\quad
\subfigure[Lighthouse]{
\includegraphics[width=0.22\linewidth]{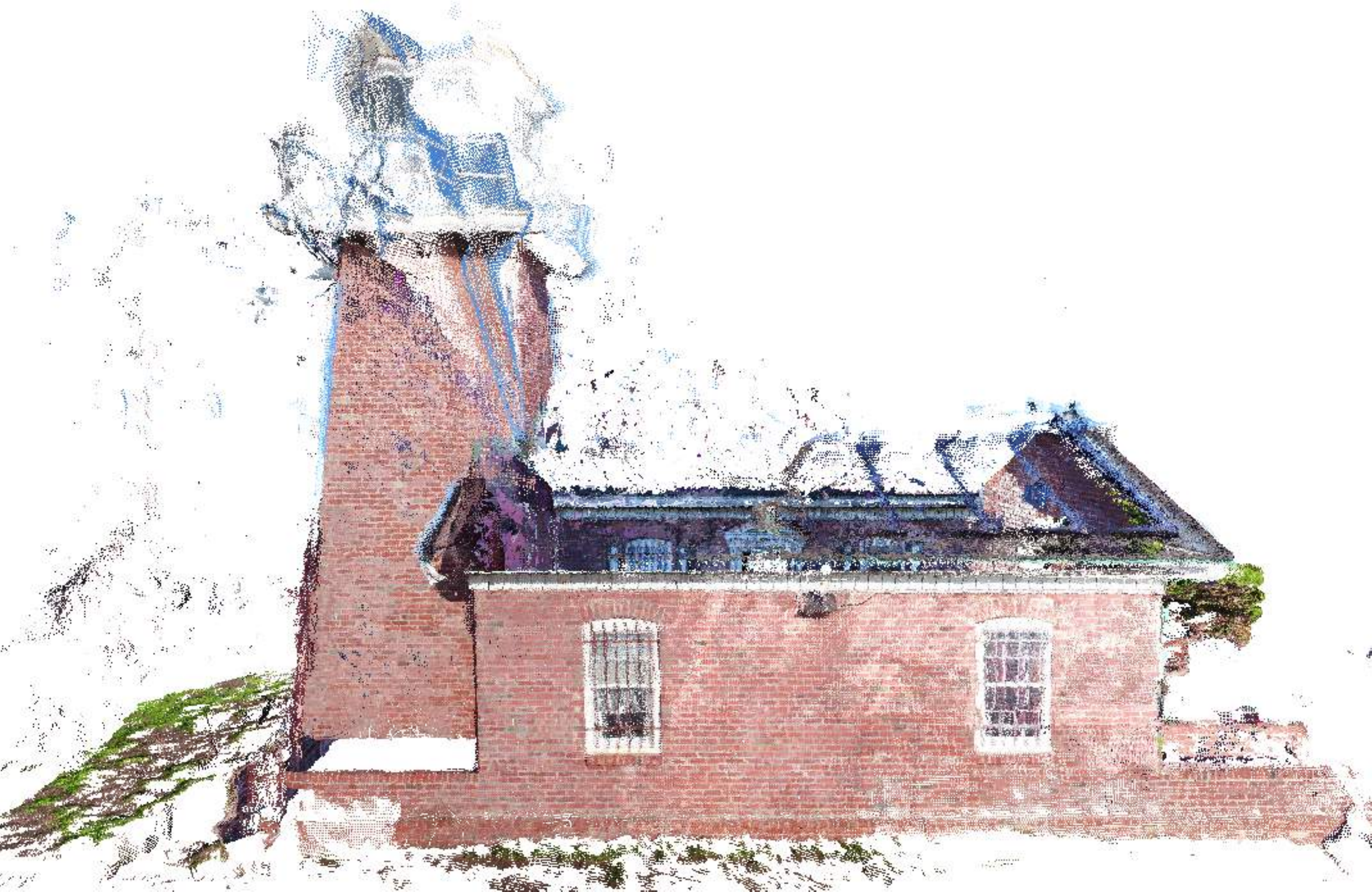}
}
\caption{The performance of M$\mathbf{^3}$VSNet on the \textsl{Tanks and Temples} benchmark \cite{knapitsch2017tanks} without any finetuning. The quality of dense point cloud reconstruction in large-scale scene shows the powerful generalization ability of M$\mathbf{^3}$VSNet.}
\label{fig:TT}
\end{figure*}

To evaluate the generalization ability of our proposed M$\mathbf{^3}$VSNet, we use the intermediate \textsl{Tanks and Temples} benchmark that has high-resolution images of outdoor large-scale scenes. The model of our proposed M$\mathbf{^3}$VSNet trained on the \textsl{DTU} dataset is transferred to the \textsl{Tanks \& Temples} benchmark without any finetuning. The intermediate \textsl{Tanks and Temples} benchmark contains kinds of images with the resolution of 1920 $\times$ 1056 and with the depth hypothesis $D=160$. Another core hyperparameter is the photometric threshold in the process of depth fusion. For the same depth maps, the different photometric thresholds will lead to different performances. Higher photometric threshold will cause better accuracy but worse completeness. In turn, lower photometric threshold will introduce better completeness but worse accuracy. For our proposed M$\mathbf{^3}$VSNet, the photometric threshold is set to 0.6 and we get the following results. As shown in table \ref{tab:TTtable}, the ranking is selected from the leaderboard of the intermediate \textsl{Tanks and Temples} benchmark. Our proposed M$\mathbf{^3}$VSNet is better than $\mathrm{MVS^2}$ by the mean score of 8 scenes, which is the best unsupervised MVS network until April 17, 2020. In table \ref{tab:TTtable}, the higher the mean score, the higher the ranking relatively. Further, the score of M$\mathbf{^3}$VSNet in Playground scene is 47.39, which is better than the score of $\mathrm{MVS^2}$ 43.48. Therefore, our proposed M$\mathbf{^3}$VSNet ranks higher than $\mathrm{MVS^2}$. As a matter of fact, the final ranking is the mean of the independent ranking of 8 scenes, which is different from the calculation method of the mean score. The dense point clouds are presented in figure \ref{fig:TT}, which are reasonable and complete for Family, Francis, Horse, M60, Panther, Playground, Train, Lighthouse scenes. Besides, the robustness of our proposed M$\mathbf{^3}$VSNet also play an important role in non-ideal areas in \textsl{Tanks and Temples} benchmark such as the sand in the scene of Playground. In view of the above, the performance in table \ref{tab:TTtable} and figure \ref{fig:TT} demonstrates the powerful generalization ability of our proposed M$\mathbf{^3}$VSNet.

\section{Conclusion}
\label{sec:conclusion}
In this paper, we propose an unsupervised multi-metric network for multi-view stereo reconstruction named M$\mathbf{^3}$VSNet, which improve the robustness and completeness of point cloud even in non-ideal environments. The proposed novel multi-metric loss function, namely pixel-wise and feature-wise loss function, can capture more semantic information to learn the inherent constraints from different perspectives of matching correspondences. The performance of point cloud reconstruction in non-ideal environments for robustness and completeness will also benefit from the multi-metric loss mainly. Besides, with the incorporation of normal-depth consistency, M$\mathbf{^3}$VSNet improves the accuracy and continuity of the estimated depth maps by the orthogonality between normal and local surface tangent. Extensive experiments show that our proposed M$\mathbf{^3}$VSNet outperforms the previous state-of-the-arts unsupervised learning-based methods and achieves comparable performance with original MVSNet \cite{yao2018mvsnet} on the \textsl{DTU} dataset and demonstrates the powerful generalization ability on the \textsl{Tanks \& Temples} benchmark \cite{knapitsch2017tanks} with effective improvement. In the future, more MVS datasets with high precision are desired. To relief the high cost of datasets, the domain transfer for different datasets can be improved and enhanced. What's more, multi-task such as object detection, semantic and instance segmentation, depth completion, etc. can be combined with multi-view stereo reconstruction for the time to come.

%%
%% The next two lines define the bibliography style to be used, and
%% the bibliography file.

\end{document}